%% file: main.tex
\documentclass[10pt,twocolumn,letterpaper]{article}

\usepackage{cvpr}              
\input{preamble}
\definecolor{cvprblue}{rgb}{0.21,0.49,0.74}
\usepackage[pagebackref,breaklinks,colorlinks,allcolors=cvprblue]{hyperref}
\usepackage{url}            
\usepackage{booktabs}       
\usepackage{amsfonts}       
\usepackage{nicefrac}       
\usepackage{microtype}      
\usepackage{xcolor}         
\usepackage{amsthm}
\usepackage{graphicx}
\usepackage{multirow}
\usepackage{array}
\usepackage{adjustbox}
\usepackage{amsfonts}
\usepackage{amsmath}
\usepackage{enumitem}
\usepackage{pifont}
\usepackage{booktabs}  
\usepackage{tabularx}  

\def\paperID{} 
\def\confName{CVPR}
\def\confYear{2026}

\title{Fast Kernel-Space Diffusion for Remote Sensing Pansharpening}

\author{
Hancong Jin\\
{\tt\small jhc261973@gmail.com}
\and
Zihan Cao\\
{\tt\small iamzihan666@gmail.com}
\and
Liang-jian Deng\\
{\tt\small liangjian.deng@uestc.edu.cn}
\and
Jingjing Li\\
{\tt\small jjl@uestc.edu.cn}
\\[1em]
University of Electronic Science and Technology of China
}

\begin{document}
\maketitle
\begin{abstract}
    Pansharpening seeks to fuse high-resolution panchromatic (PAN) and low-resolution multispectral (LRMS) images into a single image with both fine spatial and rich spectral detail. Despite progress in deep learning-based approaches, existing methods often fail to capture global priors inherent in remote sensing data distributions. Diffusion-based models have recently emerged as promising solutions due to their powerful distribution mapping capabilities, however, they suffer from heavy inference latency. We introduce KSDiff, a fast kernel-space diffusion framework that generates convolutional kernels enriched with global context to enhance pansharpening quality and accelerate inference. Specifically, KSDiff constructs these kernels through the integration of a low-rank core tensor generator and a unified factor generator, orchestrated by a structure-aware multi-head attention mechanism. We further introduce a two-stage training strategy tailored for pansharpening, facilitating integration into existing pansharpening architectures. Experiments show that KSDiff achieves superior performance compared to recent promising methods, and with over $500 \times$ faster inference than diffusion-based pansharpening baselines. Ablation studies, visualizations and further evaluations substantiate the effectiveness of our approach.
\end{abstract}

\begin{figure}[htbp]
    \centering
    \includegraphics[width=1.0\linewidth]{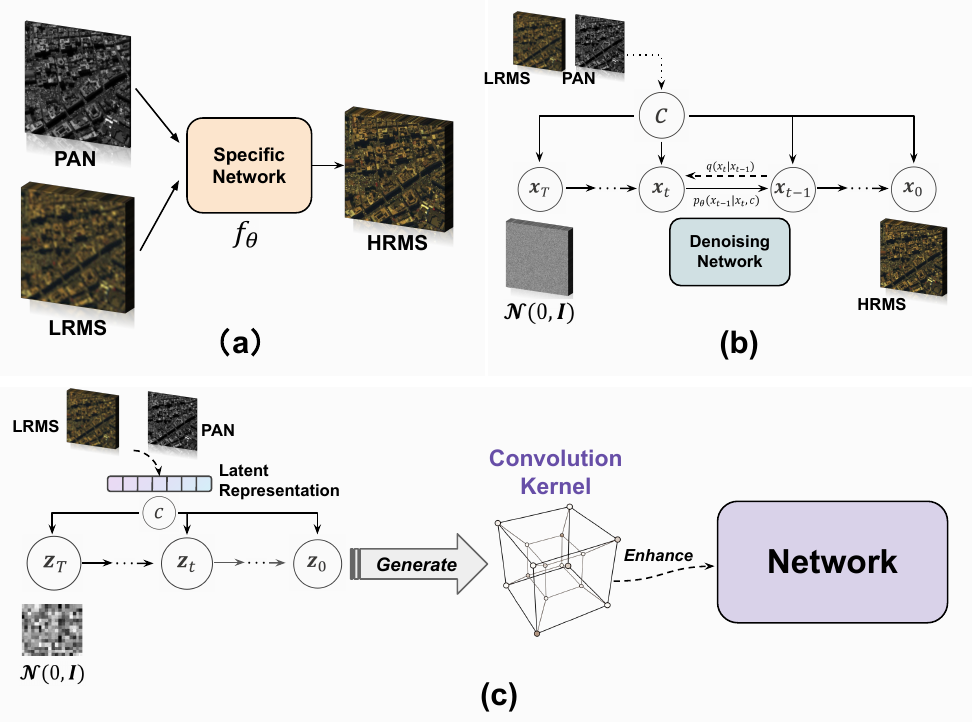} 
    \caption{
        (a) Traditional DL-based methods directly learn a non-linear mapping $f_{\theta}$ to fuse PAN and LRMS images in a one-step manner.
        (b) Recent diffusion-based methods employ a multi-step refinement process in the pixel space conditioned on PAN and LRMS from pure Gaussian noise 
            $\mathbf{\mathcal{N}}$(0, \textit{\textbf{I}}). The $q(\mathbf{x}_{t}|\mathbf{x}_{t-1})$, $p_{\theta}(\mathbf{x}_{t-1}|\mathbf{x}_{t}, \mathbf{c})$ and $\mathbf{c}$
            denote the noise adding process, the reverse denoising process, and the condition, respectively.
        (c) The proposed KSDiff, which generates convolution kernels to enhance
regression-based pansharpening networks via a diffusion model performing in the latent space. This design enables high-quality reconstruction with fast inference.
        }
    \label{fig:wwww}
\end{figure}
\section{Introduction}

Due to the inherent physical limitations of satellite sensors, the direct acquisition of 
high-resolution multispectral images (HRMS) is not feasible. Instead, sensors typically 
provide high-resolution panchromatic (PAN) images and low-resolution multispectral images 
(LRMS). Pansharpening techniques aim to fuse the spatial detail of PAN images with the 
spectral richness of LRMS, thereby generating HRMS with enhanced spatial and spectral 
fidelity. As a foundational preprocessing approach, pansharpening plays a critical role in 
numerous remote sensing applications, including image segmentation \cite{yuan2021review} and change detection \cite{wu2017post}.

Over the past decades, a variety of pansharpening techniques have been proposed, which can be 
broadly categorized into four families: component substitution (CS) \cite{kwarteng1989extracting, meng2016pansharpening},
multi-resolution analysis (MRA) \cite{otazu2005introduction, vivone2017regression}, 
variational optimization (VO) \cite{wu2023lrtcfpan, wu2023framelet}, 
and deep learning (DL)-based methods. While classical CS and MRA methods offer interpretable 
mathematical formulations, they often introduce noticeable spectral or spatial distortions due 
to oversimplified assumptions. VO-based methods improve flexibility by formulating pansharpening 
as an optimization problem with prior constraints, but their performance is sensitive to model 
design and hyperparameters.

In recent years, deep learning, particularly convolutional neural networks (CNNs), has emerged 
as a powerful paradigm for pansharpening, achieving impressive performance across benchmark datasets. 
Representative approaches such as PanNet \cite{yang2017pannet}, DiCNN \cite{he2019pansharpening}, 
and FusionNet \cite{deng2020detail} model the fusion process as a non-linear mapping from PAN and 
LRMS inputs to an HRMS output. These methods typically rely on improved architectural designs \cite{wu2021dynamic, liang2022pmacnet} 
or more effective modules \cite{jin2022lagconv, duan2024content} to enhance learning capacity. Most existing deep learning-based models are deterministic, which enables fast inference but limits their capacity to capture global statistical dependencies and the complex high-dimensional data distributions inherent in remote sensing imagery.

Diffusion models \cite{ho2020denoising, karras2022elucidating, saharia2022image}, 
rooted in generative modeling, learn complex data distributions through 
iterative denoising steps, reconstructing samples from Gaussian noise. Unlike conventional deep learning methods mentioned above
, they model the full conditional distribution of outputs given inputs, enabling 
flexible, context-aware generation. This makes them ideal for pansharpening, as they integrate 
heterogeneous inputs like PAN and LRMS images to produce spectrally consistent, high-resolution multispectral 
images with high fidelity \cite{meng2023pandiff, cao2024diffusion}. However, a common limitation of diffusion-based 
methods  lies in their slow inference speed due to the iterative nature of the sampling process \cite{song2020denoising, karras2022elucidating}. This latency issue becomes particularly critical in 
the pansharpening task, where the target images are of high resolution and the input data typically contains 
more channels than standard RGB images. These factors significantly increase the difficulty of distribution mapping within diffusion models.
Moreover, in the context of pansharpening, where most spatial and spectral information is already 
available, the role of the network is primarily to refine rather than reconstruct from scratch. 
Consequently, reconstructing the entire image from pure Gaussian noise is both counterintuitive and burdensome.

\begin{figure*}[htbp]
    \centering
    \includegraphics[width=1.0\linewidth]{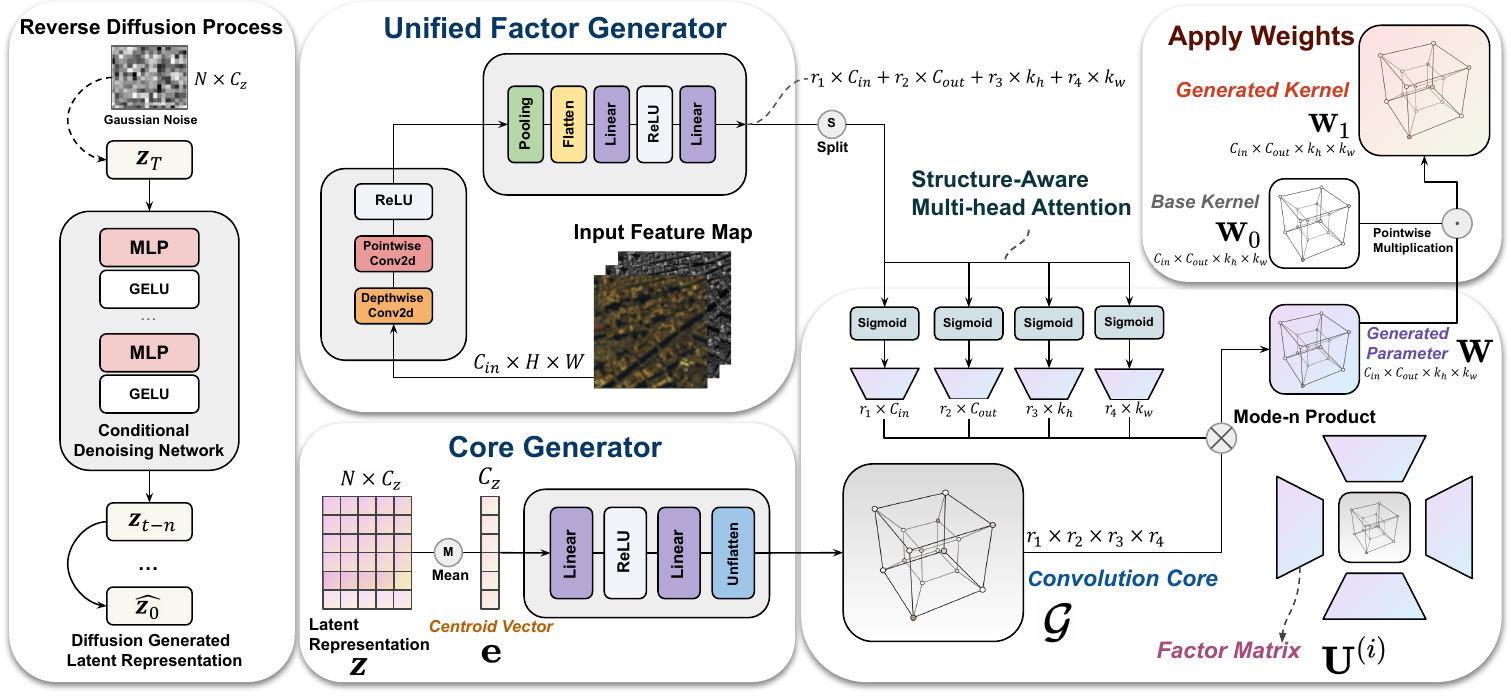}
    \caption{Kernel Generator of our proposed KSDiff. The kernel generator comprises two sub-modules: (1) a diffusion model-driven convolutional core generator, 
    (2) a unified factor generator that takes feature maps as input. The outputs of these two modules are integrated 
    using a structure-aware multi-head attention mechanism to reparameterize the base kernel. Note that in the pre-training stage, the latent representation $\mathbf{z}$ is the output of an encoder; in the diffusion model training stage and at inference time, this representation is generated by the diffusion model, namely $\hat{\mathbf{z}}_{0}$.}
    \label{fig:KG}
\end{figure*}

Considering the respective strengths and limitations of traditional deep learning methods and diffusion-based approaches 
for pansharpening, we propose a novel framework named KSDiff. KSDiff leverages the powerful distribution modeling capability of 
diffusion processes to enrich convolutional kernels with high-level remote sensing context, while preserving the inference 
efficiency of conventional DL-based methods. 
Specifically, we train a diffusion process in latent space \cite{rombach2022high, xia2023diffir}, significantly reducing the computational burden compared to modeling the diffusion process
in the full pixel space and ensuring efficient inference.  
The generated latent representations are then used to construct low-rank convolutional cores, which are subsequently expanded into full convolutional kernels 
via a structure-aware multi-head attention mechanism that integrates the generated convolution cores with features extracted from the PAN and LRMS images, as shown in \cref{fig:KG}. 
To effectively train KSDiff with regression backbones, we further introduce a two-stage training scheme and a Pyramid Latent Fusion Encoder (PLFE) tailored to the pansharpening task. 
KSDiff can be integrated into existing networks, boosting performance over baselines. Remarkably, it achieves inference speeds comparable to 
traditional DL methods, orders of magnitude faster than recent diffusion-based pansharpening approaches.  
Our main contributions are as follows:

\begin{itemize}
    \item We propose KSDiff, a novel framework that integrates diffusion-generated latent representation into the design of convolutional kernels. This approach leverages the strong distribution modeling capabilities of diffusion models while achieving fast inference speed.  
    \item We introduce a two-stage training scheme and a Pyramid Latent Fusion Encoder (PLFE) specifically tailored for pansharpening. These components enable effective integration of multi-scale spatial-spectral information and allow KSDiff to be incorporated into existing regression-based networks.
    \item We conduct extensive experiments on multiple benchmark datasets, demonstrating that KSDiff achieves competitive performance in both quantitative metrics and visual quality. In addition, it offers inference speeds comparable to traditional DL-based methods and significantly faster than recent diffusion-based pansharpening methods.
\end{itemize}

\section{Related Works}
\label{gen_inst}

\subsection{Deep Learning Based Methods}

PNN \cite{masi2016pansharpening} was the first to apply CNNs to pansharpening, inspired by single-image 
super-resolution techniques. To better preserve high-frequency details, PanNet \cite{yang2017pannet} 
introduced a strategy that injects high-pass PAN information into the upsampled LRMS image. 
DiCNN \cite{he2019pansharpening} further enhanced detail preservation using a detail-injection framework, 
while FusionNet \cite{deng2020detail} employed a residual network to explicitly learn high-frequency 
components. Subsequent works, such as DCFNet \cite{wu2021dynamic} and PMACNet \cite{liang2022pmacnet}, 
improved CNN architectures and achieved superior performance. CTINN \cite{zhou2022pan} adopted a 
transformer-based design to capture long-range dependencies in the fusion process. PanMamba\cite{he2025pan} explored the potential of the state-space model. To address the 
limitations of static convolutional kernels, LAGConv \cite{jin2022lagconv} introduced dynamic kernels 
conditioned on the input. Building on this idea, AKD \cite{peng2022source} proposed dual dynamically 
generated branches for extracting spatial and spectral details. Another branch of methods delved into 
frequency domain analysis of PAN and LRMS images, such as HFIN \cite{tan2024revisiting}.
These advancements have significantly expanded the research scope of DL-based pansharpening methods.
\subsection{Diffusion Models}
Diffusion models (DMs) \cite{ho2020denoising, song2020score} have recently advanced generative modeling by 
formulating data synthesis as a sequence of denoising steps. While initially applied to image generation 
\cite{ho2020denoising}, subsequent developments enabled high-resolution synthesis \cite{rombach2022high}, 
multimodal conditioning \cite{esser2024scaling}, and efficient sampling \cite{song2020denoising, zheng2023dpm}. 
DMs have also shown strong promise in low-level vision tasks \cite{li2025diffusion}. StableSR \cite{wang2024exploiting} achieved real-world image super-resolution
by leveraging the prior knowledge from large pre-trained text-to-image latent diffusion model. DiffIR \cite{xia2023diffir} combined the training of a
transformer with a diffusion model for realistic image restoration. DPS \cite{chung2022diffusion} introduced zero-shot posterior 
sampling for general inverse problems. In pansharpening, most methods \cite{meng2023pandiff, cao2024diffusion, rui2024unsupervised, Xiao_2025_CVPR} use conditional diffusion models \cite{saharia2022image} 
in pixel space, conditioning on PAN and LRMS to iteratively generate HRMS outputs. Recent theory \cite{albergo2023stochastic} unifies diffusion, flow matching \cite{lipman2022flow}, 
and diffusion-bridge models \cite{zhou2023denoising, de2021diffusion} under stochastic differential equations (SDEs) and ordinary differential equations (ODEs), enabling flexible 
distribution mapping. However, solving these SDEs/ODEs demands many network function evaluations (NFEs), incurring high computational cost. 
Acceleration strategies, such as advanced samplers \cite{karras2022elucidating, shaul2023bespoke} and distillation techniques \cite{liu2022flow, song2023consistency}, 
reduce NFEs but either require full model retraining or lead to performance trade-offs.
Some recent work further explores the versatile use of diffusion models. For example, Neural Network Diffusion \cite{wang2024neural} employs an autoencoder–diffusion pipeline to synthesize latent representations of network parameters and decode them into novel, high-performing weights.

\section{Methodology}
\label{headings}
In this section, we present the design of the kernel generator in our proposed KSDiff, 
as well as a two-stage training scheme tailored for the pansharpening task.
\cref{fig:KG} illustrates the architecture of the KSDiff kernel generator. \cref{fig:plfe1} shows the structure of the latent encoder.
\cref{fig:pipe} depicts the two-stage training pipeline and the inference. We introduce our method following the order of 
training procedures.

\subsection{Pre-training Stage}
\begin{figure}[htbp]
    \centering
    \includegraphics[width=1.0\linewidth]{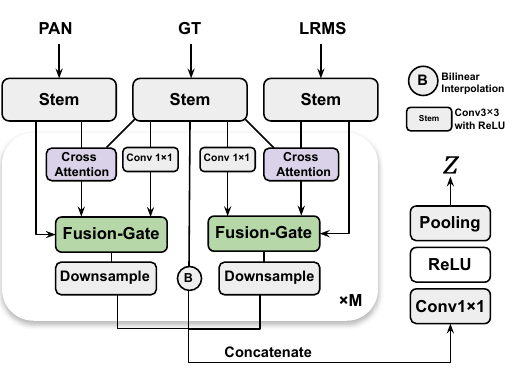} 
    \caption{
        Pyramid Latent Fusion Encoder (PLFE). The figure only shows the structure of $\mathrm{PLFE}_1$; since $\mathrm{PLFE}_2$ only takes PAN and LRMS as input, its structure is slightly different (a halved version compared to $\mathrm{PLFE}_1$). \textit{Further details can be found in supplementary materials \cref{sec: further methods}}.
        }
    \label{fig:plfe1}
\end{figure}

\begin{figure*}[htbp]
    \centering
    \includegraphics[width=1.0\linewidth]{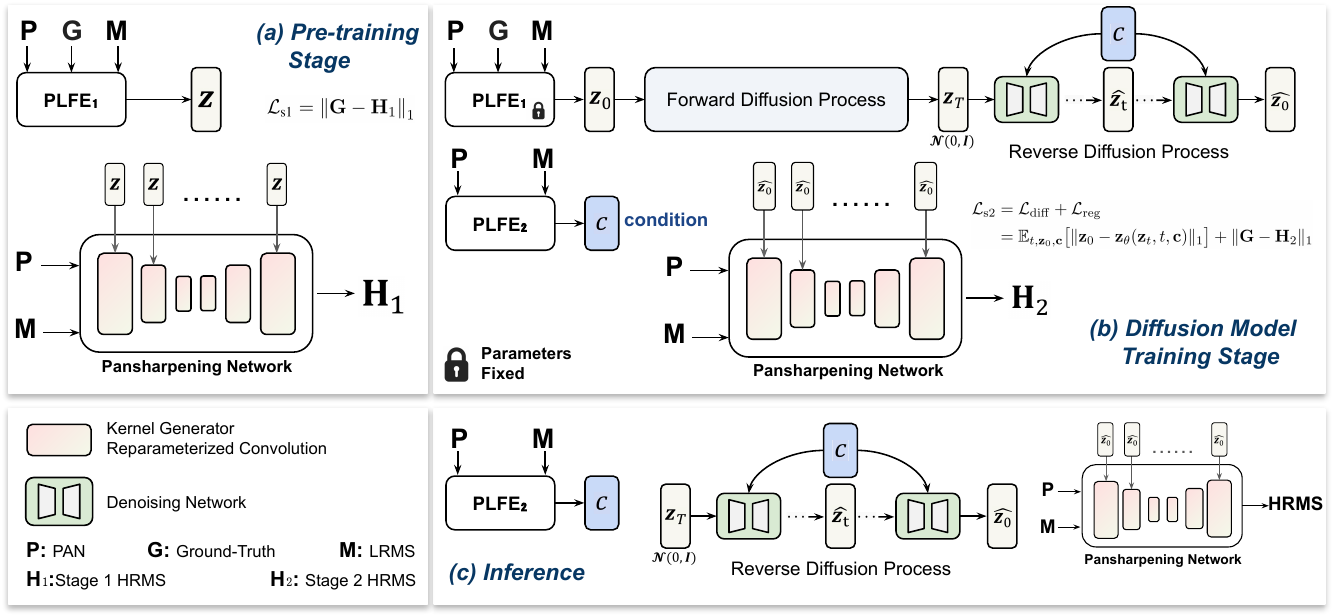}
    \caption{An overview of our two-stage training procedure and inference process.
(a) \textbf{Pre-training Stage}: The goal is to extract a latent representation $\mathbf{z}$ by optimizing $\mathrm{PLFE}_1$ jointly with the kernel generator and the pansharpening network.
(b) \textbf{Diffusion Model Training Stage}: The latent representation $\mathbf{z}_0$, extracted by the pre-trained $\mathrm{PLFE}_1$, is predicted by leveraging the strong distribution estimation ability of the diffusion model (no ground truth is used as input for its denoising network).
(c) \textbf{Inference}: Only $\mathrm{PLFE}_2$, the reverse diffusion process, the kernel generator, and the pansharpening network are involved.
            } 
    \label{fig:pipe}
\end{figure*}

In the pre-training stage, our objective is to encode the ground truth high-resolution multispectral (HRMS) image into a 
compact latent representation, which is then utilized to guide the convolutional kernels within pansharpening networks 
during image fusion. As illustrated in \cref{fig:plfe1} and \cref{fig:pipe}, we employ a Pyramid Latent Fusion Encoder (PLFE) 
to compress the ground truth HRMS image and extract a compact prior feature. This prior feature is subsequently used to generate low-rank 
convolutional core tensors (\textit{i.e.}, $\mathcal{G}$ in Fig. \ref{fig:KG}), from which full convolutional kernels are reconstructed. By incorporating this learned prior, our method enables 
kernel-wise guidance across the pansharpening network, thereby enhancing the fidelity and spatial-spectral consistency of the reconstructed multispectral image. 
The components of this framework are described in detail below.

\paragraph{Pyramid Latent Fusion Encoder (PLFE).}  
To fully leverage the complementary priors of the PAN, LRMS and HRMS modalities while avoiding the spatial–spectral entanglement inherent in naive concatenation, we introduce the Pyramid Latent Fusion Encoder (PLFE). Its design is driven by two principles: first, a multi-scale pyramid architecture in which PAN and LRMS features are progressively refined under HRMS guidance at each level, integrating high-resolution spatial cues with contextual spectral semantics; and second, a dynamic fusion gate that adaptively balances the contributions of the original branch features and the HRMS-guided features to maintain spatial–spectral consistency throughout the encoding process.
Specifically, given the PAN image $\mathbf{P}\in\mathbb{R}^{H\times W\times1}$, 
the LRMS image $\mathbf{M}\in\mathbb{R}^{H\times W\times C}$ and 
the ground-truth HRMS image $\mathbf{G}\in\mathbb{R}^{H\times W\times C}$, 
PLFE first embeds each via a stem block, followed by $M$ pyramid stages to capture features at multi-scale.
At a certain stage, the PAN or LRMS branch features $\mathbf{X} \in \mathbb{R}^{H'\times W'\times{d}}$ are refined
under guidance from the HRMS feature $\mathbf{Y} \in \mathbb{R}^{H'\times W'\times{d}}$ using cross attention with linear complexity \cite{shen2021efficient, katharopoulos2020transformers}, alleviating the 
computational burden of this dual-branch attention when facing large images. Its mechanism (\textit{see supplementary materials \cref{fig:plfe2_ca} (b) for graphic illustration}) is as follows:
\begin{equation}
\begin{aligned}
\mathbf{Q} &= \mathrm{Rsp}(\mathrm{Softmax}(\mathbf{Q}, 1)), \mathbf{K} = \mathrm{Rsp}(\mathrm{Softmax}(\mathbf{K}, 2)) \\
\mathbf{V} &= \mathrm{Rsp}(\mathbf{V}), \mathbf{A} = \mathbf{K} \odot \mathbf{V}^{T},\mathbf{O} = \mathrm{Conv}(\mathbf{A}^{T} \odot \mathbf{Q})
\end{aligned}
\end{equation}
where $\mathbf{Q}$, $\mathbf{K}$, $\mathbf{V}$, $\mathbf{A}$, $\mathbf{O}$ represent the query, key, value, attention map, and output, respectively, $\odot$ is the elementwise multiplication.
$\mathbf{Q}$, $\mathbf{K}$ and $\mathbf{V}$ are obtained from $\mathbf{X}$ and $\mathbf{Y}$ with several convolution layers.
$\mathrm{Rsp}$ (Reshape) operation flattens the spatial dimensions of $\mathbf{Q}$, $\mathbf{K}$, and $\mathbf{V}$, into 
a single dimension, i.e., $\mathbb{R}^{H \times W \times d} \to \mathbb{R}^{HW \times d}$, 
and $\mathrm{Softmax}(\mathbf{K}, j)$ represents the $\mathrm{Softmax}$ operation along the $j$th dimension.
Consequently, the memory complexity is reduced from $\mathcal{O}((HW)^{2})$ to $\mathcal{O}(d^{2})$, with $d \ll HW$, compared to vanilla attention \cite{vaswani2017attention}. 
We then adaptively fuse the guidance from the cross attention via a dynamic fusion gate:  
\begin{equation}
\begin{aligned}
&\mathbf{G}_{\mathrm{gate}} = 
    \sigma\bigl(\mathrm{Conv}_g\bigl[\mathbf{X},\,\mathrm{Proj}(\mathbf{Y})\bigr]\bigr), \\
&\mathbf{F} = \mathbf{G}_{\mathrm{gate}}\odot\mathbf{X} 
    + (1-\mathbf{G}_{\mathrm{gate}})\odot\mathrm{Proj}(\mathbf{Y}) 
    + \mathbf{O}.
\end{aligned}
\end{equation}
where $\mathrm{Proj}$ is for channel alignment, $\bigl[,\bigr]$ means concatenation, and $\sigma$ is a $\mathrm{Sigmoid}$ function along channel dimension. 
By learning $\mathbf{G}_{\mathrm{gate}}$, the network balances trusting the HRMS prior-refining features where reliable, 
and preserving the original branch feature where guidance may misalign, thus enhancing spatial-spectral consistency and reducing artifacts, yielding better encoding quality. 
A strided convolution downsamples $\mathbf{F}$, while $\mathbf{Y}$ is bilinearly interpolated for the next stage. After $M$ stages, 
the final features are concatenated and then projected into $\mathbf{z}\in\mathbb{R}^{N\times C_z}$, with $N$ and $C_z$ representing the token number 
and the embedded dimension respectively, obtaining a compact prior representation ($N \ll HW$) for downstream kernel modulation.  

\paragraph{Kernel Generator.}

We explore an efficient strategy to enrich convolutional layers with encoded latent representations 
that capture high-level global context. Given a convolutional kernel represented as a 4D tensor of 
shape $C_{\text{in}} \times C_{\text{out}} \times k_h \times k_w$ (typically with $k_h = k_w = k$), 
a naive solution is to flatten the latent code $\mathbf{z}$, pass it through an MLP, and reshape 
the output to obtain the convolution kernel. However, this approach introduces a prohibitive number 
of parameters and lacks flexibility in controlling the influence of the latent representation across different kernels within a network. 
To address these issues, we propose to modulate a standard base kernel 
$\mathbf{W}_0 \in \mathbb{R}^{C_{\text{in}} \times C_{\text{out}} \times k_h \times k_w}$ 
with a learned weight tensor $\mathbf{W}$, yielding the final kernel as 
$\mathbf{W}_1 = \mathbf{W}_0 \odot \mathbf{W}$, where $\odot$ denotes element-wise multiplication. 
To reduce complexity, we factorize $\mathbf{W}$ via tensor algebra \cite{tucker1966some, kolda2009tensor}: 
\begin{equation}
\mathbf{W} = \mathcal{G} \times_1 \mathbf{U}^{(1)} \times_2 \mathbf{U}^{(2)} \times_3 \mathbf{U}^{(3)} \times_4 \mathbf{U}^{(4)},
\end{equation}
where $\times_n$ denotes mode-$n$ product, with a compact core tensor $\mathcal{G} \in \mathbb{R}^{r_1 \times r_2 \times r_3 \times r_4}$ 
and factor matrices $\{\mathbf{U}^{(n)}\}$ of 
size $\mathbf{U}^{(1)} \in \mathbb{R}^{r_1 \times C_{\text{in}}}$, $\mathbf{U}^{(2)} \in \mathbb{R}^{r_2 \times C_{\text{out}}}$, $\mathbf{U}^{(3)} \in \mathbb{R}^{r_3 \times k_h}$, and $\mathbf{U}^{(4)} \in \mathbb{R}^{r_4 \times k_w}$. 
The core tensor is generated by applying mean pooling on $\mathbf{z}$ to obtain a centroid vector $\mathbf{e} \in \mathbb{R}^{C_z}$, which is processed by an MLP. 
The factor matrices are obtained from input features via a lightweight shared backbone followed by four attention heads, 
which we refer to as structure-aware multi-head attention, as shown in \cref{fig:KG}. 
This design achieves efficient, expressive kernel modulation with controllable prior influence. An experiment on the influence of different low-rank core tensor sizes is presented in \cref{discuss}.
In comparison to the naive MLP-based approach with $\mathcal{O}(C_{\text{in}} C_{\text{out}} k^2 C_z)$ complexity, 
this formulation has $\mathcal{O}(C_z r_1 r_2 r_3 r_4 + \sum_{n=1}^{4} r_n d_n)$ theoretical complexity, 
where $d_n$ denotes the size of the $n$-th mode. With $r_n \ll d_n$, this leads to notable savings in computation and memory while supporting flexible 
kernel-wise adaptation.

\paragraph{Pansharpening Network.}
To better demonstrate the performance gains achieved by our proposed KSDiff, we construct a typical U-Net~\cite{ronneberger2015u} as the pansharpening backbone and integrate our kernel generation method into several of its convolutional layers. \textit{A detailed description of the network architecture is provided in supplementary materials \cref{sec: further methods}.}

\paragraph{Training Technique.}
To enhance the capability of the latent encoder in constructing an informative prior feature representation, we jointly optimize it alongside the kernel generator within the integrated pansharpening framework. The training objective employs an $L_1$ loss function, which is commonly chosen by reconstruction tasks \cite{cao2024diffusion, yang2019deep},  formulated as:
\begin{equation}
\mathcal{L}_{\text{s1}} = \left \| \mathbf{G} - \mathbf{H}_{1} \right \|_{1},
\end{equation}
where $\mathbf{G}$ denotes the ground-truth high-resolution multispectral (HRMS) image, and $\mathbf{H}_{1}$ represents the reconstructed HRMS output.

\subsection{Diffusion Model Training Stage}
In this stage, as illustrated in \cref{fig:pipe} (b), 
we leverage the powerful distribution estimation capability 
of diffusion models to approximate the latent representation prior. 
Our framework adopts both DDPM \cite{ho2020denoising} and DDIM \cite{song2020denoising} paradigms to formulate the forward and 
backward processes. While more recent advancements in diffusion scheduling \cite{karras2022elucidating, lipman2022flow} or 
more efficient sampling strategies \cite{zheng2023dpm, lu2022dpm, yue2023resshift} could potentially be incorporated into our framework, 
we intentionally employ the classical approach to maintain both simplicity and excellent 
pansharpening performance.

\paragraph{Diffusion Model.}
In the forward diffusion process, given the ground-truth high-resolution multispectral (HRMS) image, we first employ the pre-trained $\mathrm{PLFE}_1$ from the previous stage to obtain the corresponding prior feature $\mathbf{z}_{0} \in \mathbb{R}^{N \times C_{z}}$, which serves as the initial state of the forward Markov chain. We then gradually add Gaussian noise over $T$ iterations as follows:
\begin{equation}
    \begin{aligned}
     q(\mathbf{z}_{1:T}\mid \mathbf{z}_{0})
    &= \prod_{t=1}^{T}q(\mathbf{z}_{t}\mid \mathbf{z}_{t-1}),
    \\
    q(\mathbf{z}_{t}\mid \mathbf{z}_{t-1})
    &= \mathcal{N}\bigl(\mathbf{z}_{t};\sqrt{1-\beta_{t}}\,\mathbf{z}_{t-1},\;\beta_{t}\mathbf{I}\bigr),   
    \end{aligned}
\end{equation}
where $t=1,\ldots,T$, $\mathbf{z}_{t}$ denotes the noisy feature at step $t$, and $\beta_{1:T}\in(0,1)$ controls the noise schedule. By utilizing the reparameterization trick \cite{kingma2013auto}, one can derive the closed-form marginal:
\begin{equation}
    \begin{aligned}
     &q(\mathbf{z}_{t}\mid \mathbf{z}_{0})
    = \mathcal{N}\bigl(\mathbf{z}_{t};\sqrt{\bar{\alpha}_{t}}\,\mathbf{z}_{0},\;(1-\bar{\alpha}_{t})\mathbf{I}\bigr),
    \\
    &\alpha_{t}=1-\beta_{t}, \quad
    \bar{\alpha}_{t}=\prod_{i=1}^{t}\alpha_{i}.   
    \end{aligned}
\end{equation}
The reverse diffusion process aims to recover the prior feature from an isotropic Gaussian initialization. It is formulated as a $T$-step Markov chain running backward from $\mathbf{z}_{T}$ to $\mathbf{z}_{0}$. At each reverse step, the posterior distribution is given by:
\begin{equation}
    \begin{aligned}
&p(\mathbf{z}_{t-1}\mid \mathbf{z}_{t}, \mathbf{z}_{0})
= \mathcal{N}\!\bigl(\mathbf{z}_{t-1};\, \boldsymbol{\mu}_{t}, \boldsymbol{\Sigma}_{t}\bigr), \\[4pt]
&\boldsymbol{\mu}_{t}
= \tfrac{1}{\sqrt{\alpha_{t}}}
   \left(\mathbf{z}_{t}
   - \tfrac{1-\alpha_{t}}{\sqrt{1-\bar{\alpha}_{t}}}\,
     \boldsymbol{\epsilon}\right), \quad \boldsymbol{\Sigma}_{t}
= \tfrac{1-\bar{\alpha}_{t-1}}{1-\bar{\alpha}_{t}}\,\beta_{t}\,\mathbf{I}.
\end{aligned}
\end{equation}
where $\boldsymbol{\epsilon}$ denotes the noise component in $\mathbf{z}_{t}$, which can be estimated by a denoising network $\boldsymbol{\epsilon}_{\theta}$ (see \cref{fig:KG} and \cref{fig:pipe} (b)).  
Since we perform diffusion in the latent space, we adopt a second PLFE, denoted $\mathrm{PLFE}_{2}$, to condense the panchromatic image $\mathbf{P}$ and the LRMS image $\mathbf{M}$ into the conditional latent $\mathbf{c}\in\mathbb{R}^{N\times C_{z}}$. Note that $\mathrm{PLFE}_{2}$ retains a halved structure compared to the pre-trained $\mathrm{PLFE}_{1}$, which is symmetric(see \cref{fig:plfe1}). To accelerate sampling, we employ the DDIM sampler \cite{song2020denoising}, which operates in a non-Markovian manner:
\begin{equation}
    \begin{aligned}
    \mathbf{z}_{t-n}
    &= \sqrt{\bar{\alpha}_{t-n}}\,
       \frac{\mathbf{z}_{t}-\sqrt{1-\bar{\alpha}_{t}}\,
             \boldsymbol{\epsilon}_{\theta}(\mathbf{z}_{t},t,\mathbf{c})}
            {\sqrt{\bar{\alpha}_{t}}} \\
    &\quad + \sqrt{1-\bar{\alpha}_{t-n}-\sigma_{t}^{2}}\;
       \boldsymbol{\epsilon}_{\theta}(\mathbf{z}_{t},t,\mathbf{c})
       + \sigma_{t}^{2}\,\boldsymbol{\epsilon}.
    \end{aligned}
\end{equation}
where $\sigma_{t}$ is a predetermined function of $t$ and $n$ denotes the number of skipped steps. After sampling, we obtain the predicted prior $\hat{\mathbf{z}}_{0}\in\mathbb{R}^{N\times C_{z}}$, which is then fed into the kernel generation process to guide the entire pansharpening network.

\paragraph{Training Technique.}
The training objective of standard DDPM (denoted as \(\boldsymbol\epsilon\)-prediction) often underperforms in some specific tasks \cite{liu2024residual, cao2024diffusion}. We therefore reparameterize the diffusion loss to predict the original sample \(\mathbf{z}_0\), which is mathematically equivalent to \(\boldsymbol\epsilon\)-prediction \cite{ho2020denoising}. Concretely, we optimize:
\begin{equation}
\mathcal{L}_{\mathrm{diff}}
= \mathbb{E}_{t,\mathbf{z}_0,\mathbf{c}}\bigl[\|\mathbf{z}_0 - \mathbf{z}_\theta(\mathbf{z}_t, t, \mathbf{c})\|_1\bigr].
\end{equation}
We observe that jointly training the diffusion model and the pansharpening regressor yields superior high-resolution multispectral (HRMS) reconstructions compared to a separate training scheme, corroborating findings in other domains \cite{he2023hqg, xia2023diffir}. Accordingly, our overall objective becomes:
\begin{equation}
    \begin{aligned}
    \mathcal{L}_{\mathrm{s2}}
&= \mathcal{L}_{\mathrm{diff}} + \lambda\mathcal{L}_{\mathrm{reg}} \\
&= \mathbb{E}_{t,\mathbf{z}_0,\mathbf{c}}\bigl[\|\mathbf{z}_0 - \mathbf{z}_\theta(\mathbf{z}_t, t, \mathbf{c})\|_1\bigr]
+ \lambda\|\mathbf{G} - \mathbf{H}_2\|_1
    \end{aligned}
\end{equation}
where \(\mathbf{G}\) is the ground-truth HRMS image, and \(\mathbf{H}_2\) denotes the pansharpened output produced during diffusion-model training. The weighting factor $\lambda$ is empirically set to 1. This unified loss encourages both accurate diffusion estimation and faithful spectral-spatial reconstruction in a single end-to-end framework.

\subsection{Inference.}
At inference time (see \cref{fig:pipe} (c)), only the reverse diffusion process is employed. The $\mathrm{PLFE}_{2}$ module extracts a conditional code $\mathbf{c}$ from PAN and LRMS images, which guides the sampling process starting from pure Gaussian noise $\mathbf{z}_{T} \in \mathbb{R}^{N \times C_{z}}$ through the trained conditional denoising network. The resulting latent representation $\hat{\mathbf{z}}_{0}$ is then fed into the kernel generator, enabling the pansharpening network to fuse PAN and LRMS images with the global context provided by this representation.

\begin{figure*}[htbp]
    \centering
    \includegraphics[width=1.0\linewidth]{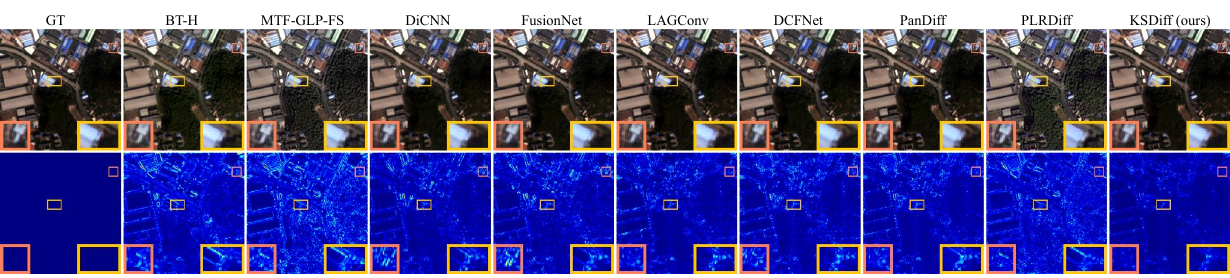}
    \caption{Comparison of qualitative results for representative methods on the GF2 reduced-resolution dataset. The first row displays RGB outputs, and the second row presents the error maps.}
    \label{fig:compare}
\end{figure*}

\section{Experiments}

\begin{table*}[htbp]
    \centering
    \small
    \caption{Result on the WV3 reduced-resolution and full-resolution datasets. The best results are highlighted in bold and the second best results are underlined.}
    \begin{adjustbox}{max width=\textwidth}
    \begin{tabular}{l|cccc|ccc|cccc}
    \toprule
    \multirow{2}{*}{Method} & \multicolumn{4}{c|}{\textbf{Reduced resolution}} & \multicolumn{3}{c|}{\textbf{Full resolution}} & \multicolumn{1}{c}{\textbf{Runtime (s)}}\\
    & SAM $\downarrow$ ($\pm$ std) & ERGAS $\downarrow$ ($\pm$ std) & Q2\textsuperscript{n} $\uparrow$ ($\pm$ std) & SCC $\uparrow$ ($\pm$ std) & $D_\lambda$ $\downarrow$ ($\pm$ std) & $D_s$ $\downarrow$ ($\pm$ std) & HQNR $\uparrow$ ($\pm$ std) &\\
    \midrule
    BDSD-PC \cite{vivone2019robust} & 5.4675$\pm$0.3842 & 4.6549$\pm$0.3278 & 0.8117$\pm$0.0238 & 0.9049$\pm$0.0094 & 0.0625$\pm$0.0053 & 0.0730$\pm$0.0080 & 0.8698$\pm$0.0119 & 0.059\\
    MTF-GLP-FS \cite{vivone2018full} & 5.3233$\pm$0.3700 & 4.6452$\pm$0.3229 & 0.8177$\pm$0.0227 & 0.8984$\pm$0.0104 & \underline{0.0206$\pm$0.0018} & 0.0630$\pm$0.0064 & 0.9180$\pm$0.0077 & 0.023\\
    BT-H \cite{aiazzi2006mtf} & 4.8985$\pm$0.2913 & 4.5150$\pm$0.2977 & 0.8182$\pm$0.0228 & 0.9240$\pm$0.0054 & 0.0574$\pm$0.0052 & 0.0810$\pm$0.0084 & 0.8670$\pm$0.0121 & 0.321\\
    PNN \cite{masi2016pansharpening} & 3.6798$\pm$0.1705 & 2.6819$\pm$0.1448 & 0.8929$\pm$0.0206 & 0.9761$\pm$0.0017 & 0.0213$\pm$0.0018 & 0.0428$\pm$0.0033 & 0.9369$\pm$0.0047 & 0.042\\
    DiCNN \cite{he2019pansharpening} & 3.5929$\pm$0.1705 & 2.6733$\pm$0.1482 & 0.9004$\pm$0.0195 & 0.9763$\pm$0.0016 & 0.0362$\pm$0.0025 & 0.0426$\pm$0.0039 & 0.9195$\pm$0.0058 & 0.083\\
    MSDCNN \cite{wei2017multi} & 3.7773$\pm$0.1796 & 2.7608$\pm$0.1539 & 0.8900$\pm$0.0201 & 0.9741$\pm$0.0017 & 0.0230$\pm$0.0020 & 0.0467$\pm$0.0045 & 0.9316$\pm$0.0061 & 0.112\\
    FusionNet \cite{deng2020detail} & 3.3252$\pm$0.1560 & 2.4666$\pm$0.1442 & 0.9044$\pm$0.0202 & 0.9807$\pm$0.0015 & 0.0239$\pm$0.0020 & 0.0364$\pm$0.0031 & \underline{0.9406$\pm$0.0044} & 0.065\\
    CTINN \cite{zhou2022pan} & 3.2523$\pm$0.1439 & 2.3936$\pm$0.1161 & 0.9056$\pm$0.0188 & 0.9826$\pm$0.0014 & 0.0550$\pm$0.0064 & 0.0679$\pm$0.0070 & 0.8815$\pm$0.0109 & 1.329\\
    LAGConv \cite{jin2022lagconv} & 3.1042$\pm$0.1249 & 2.2999$\pm$0.1368 & 0.9098$\pm$0.0188 & 0.9830$\pm$0.0015 & 0.0368$\pm$0.0033 & 0.0418$\pm$0.0036 & 0.9230$\pm$0.0055 & 1.381\\
    MMNet \cite{zhou2023memory} & 3.0844$\pm$0.1430 & 2.3428$\pm$0.1400 & 0.9155$\pm$0.0191 & 0.9829$\pm$0.0013 & 0.0540$\pm$0.0052 & \underline{0.0336$\pm$0.0026} & 0.9143$\pm$0.0063 & 0.348\\
    DCFNet \cite{wu2021dynamic} & 3.0264$\pm$0.1653 & \underline{2.1588$\pm$0.1020} & 0.9051$\pm$0.0197 & \underline{0.9861$\pm$0.0009} & 0.0781$\pm$0.0182 & 0.0508$\pm$0.0076 & 0.8771$\pm$0.0225 & 0.548\\
    PanMamba \cite{he2025pan} & \underline{2.9132$\pm$0.1230} & 2.1843$\pm$0.1092 & \underline{0.9204$\pm$0.0187} & 0.9855$\pm$0.0011 & \textbf{0.0183$\pm$0.0018} & 0.0531$\pm$0.0078 & 0.9304$\pm$0.0061 & 0.405\\
    PanDiff \cite{meng2023pandiff} & 3.2968$\pm$0.2240 & 2.4647$\pm$0.1305 & 0.8935$\pm$0.0193 & 0.9860$\pm$0.0014 & 0.0273$\pm$0.0027 & 0.0542$\pm$0.0059 & 0.9203$\pm$0.0081 & 261.410\\
    PLRDiff \cite{rui2024unsupervised} & 4.3704$\pm$0.3288 & 3.4408$\pm$0.2627 & 0.8539$\pm$0.0297 & 0.9215$\pm$0.0081 & 0.1796$\pm$0.0083 & 0.1037$\pm$0.0060 & 0.7361$\pm$0.0113 & 40.142\\
    KSDiff (ours) & \textbf{2.8102$\pm$0.1147} & \textbf{2.0756$\pm$0.0973} & \textbf{0.9221$\pm$0.0183} & \textbf{0.9870$\pm$0.0010} & 0.0210$\pm$0.0019 & \textbf{0.0322$\pm$0.0026} & \textbf{0.9468$\pm$0.0043} & 0.077\\
    \bottomrule
    \end{tabular}
    \end{adjustbox}
    \label{wv3}
\end{table*}

\begin{figure*}[htbp]
    \centering
    \includegraphics[width=1.0\linewidth]{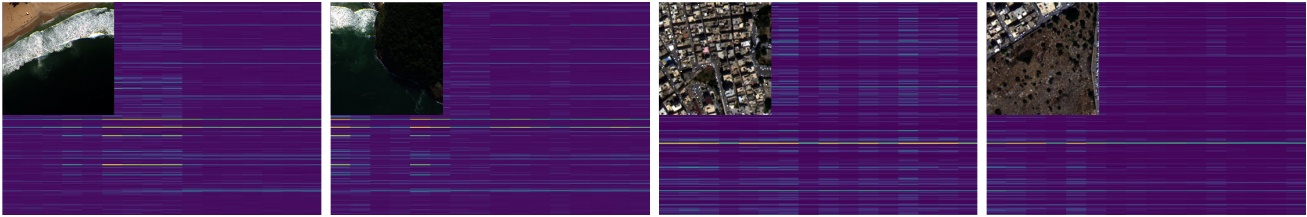}
    \caption{Visualization of latent representations generated by KSDiff from PAN and LRMS images across different scenes.}
    \label{fig:vis_latent}
\end{figure*}
    \subsection{Datasets, Metrics and Implementation Details}
    We validate our approach using datasets constructed following Wald's protocol \cite{deng2020detail, wald1997fusion}, 
    based on satellite data from WorldView-3 (WV3), GaoFen-2 (GF2), and QuickBird (QB). The data and preprocessing methods 
    are obtained from the PanCollection repository \cite{deng2022machine}. 
    We evaluate our method using SAM \cite{boardman1993automating}, ERGAS \cite{wald2002data}, Q2\textsuperscript{n} \cite{garzelli2009hypercomplex}, and SCC \cite{zhou1998wavelet} 
    for reduced-resolution datasets, and HQNR \cite{arienzo2022full}, $D_\lambda$, and $D_s$ for full-resolution datasets. 
    The model is trained with the AdamW optimizer \cite{loshchilov2017decoupled} on a single NVIDIA GeForce RTX 4090 GPU.
   Additional details on datasets, compared methods, and training configurations are provided in \textit{supplementary materials \cref{sec: exp details}}, including those related to the main experiments, ablation study, and discussion section.
    \subsection{Experimental Results}
   Our proposed KSDiff demonstrates superior performance through extensive evaluations on three benchmark datasets: WV3, GF2, and QB. \cref{wv3} to \cref{qb} compare KSDiff with state-of-the-art methods, including traditional, conventional DL-based, and diffusion-based approaches, confirming its robustness and ability to produce high-quality pan-sharpened images with fast inference. Visual comparisons in \cref{fig:compare} show KSDiff's results closely align with the ground truth. 

    \subsection{Ablation Study}
    \label{ablation}
    To validate the effectiveness of each component in our proposed method, we conducted ablation studies, with results summarized in \cref{ablation}.
    Firstly, to assess the impact of the latent diffusion prior, we evaluated our pansharpening network without incorporating the latent representation generated by the diffusion model. The results demonstrate that omitting the latent representation compromises performance, despite a reduction in latency.
    Secondly, we ablated the proposed Pyramid Latent Fusion Encoder (PLFE) by replacing it with an encoder that directly concatenates the PAN, LRMS, and GT images as input. This substitution led to degraded performance, highlighting the importance of PLFE in our framework.
    We further investigated the tensor structure-aware kernel generator by replacing it with an MLP containing an equivalent number of intermediate features. This modification increased the number of learnable parameters by over tenfold, resulting in convergence failure, underscoring the efficiency and necessity of our tensor-aware design.
    Finally, to evaluate the joint training strategy in the diffusion model training stage, we conducted an ablation by first training the diffusion model to learn the latent representation independently and then integrating it with a pre-trained regression network for testing. This approach yielded inferior performance compared to our joint training strategy, confirming its critical role in achieving optimal results.

    \subsection{Discussions}
    \label{discuss}

        \paragraph{Latent Representation Visualization.}
        As shown in \cref{fig:vis_latent}, we visualize latent diffusion representations from PAN and LRMS image pairs across diverse scenes. Representations from ocean-dominated images (left two subfigures) exhibit similar distributions, as do those from building and land-dominated images (right two subfigures), with variations due to building density. This demonstrates the diffusion model's ability to effectively capture intrinsic global information in PAN and LRMS images.
    
        \paragraph{Other Backbones.}
        To further assess the effectiveness of KSDiff in enhancing pansharpening networks, we substituted convolutional layers in various pansharpening backbones with KSDiff. The results, shown in \cref{replace}, indicate that KSDiff consistently improves the performance of DiCNN \cite{he2019pansharpening}, FusionNet \cite{deng2020detail}, and LAGNet \cite{jin2022lagconv}.
        
        \paragraph{Size of Low-Rank Core Tensor.} We investigated how the size of the low-rank core tensor $\mathcal{G} \in \mathbb{R}^{r_1 \times r_2 \times r_3 \times r_4}$ affects performance on the WV3 reduced-resolution dataset by scaling its dimensions. Using FusionNet \cite{deng2020detail} with our KSDiff-enhanced backbone, we found that smaller core tensors can achieve comparable or even better results than larger ones under the same conditions, similar to LoRA \cite{hu2022lora}. However, as shown in \cref{core}, increasing the last two modes (kernel size) from 1 to 2 improves performance, likely because a $(r_1, r_2, 1, 1)$ tensor collapses to a matrix, losing useful 4D structure. This indicates that maintaining higher-dimensional tensor structure can benefit model performance.
\begin{table}[htbp]
    \centering
    \small
    \begin{minipage}[t]{0.48\textwidth}
        \centering
        \caption{Result on the GF2 reduced-resolution dataset. The best results are highlighted in bold and the second best results are underlined.}
        \begin{adjustbox}{max width=\linewidth}
        \begin{tabular}{l|cccccccccc}
        \toprule
        \multirow{2}{*}{Method} & \multicolumn{4}{c}{\textbf{GaoFen-2}} \\
        & SAM $\downarrow$ ($\pm$ std) & ERGAS $\downarrow$ ($\pm$ std) & Q2\textsuperscript{n} $\uparrow$ ($\pm$ std) & SCC $\uparrow$ ($\pm$ std) \\
        \midrule
        BDSD-PC \cite{vivone2019robust} & 1.7110$\pm$0.0718 & 1.7025$\pm$0.0907 & 0.8932$\pm$0.0069 & 0.9448$\pm$0.0037 & \\
        MTF-GLP-FS \cite{vivone2018full} & 1.6757$\pm$0.0773 & 1.6023$\pm$0.0793 & 0.8914$\pm$0.0057 & 0.9390$\pm$0.0044 & \\
        BT-H \cite{aiazzi2006mtf} & 1.6810$\pm$0.0709 & 1.5524$\pm$0.0814 & 0.9089$\pm$0.0065 & 0.9508$\pm$0.0034 & \\
        PNN \cite{masi2016pansharpening} & 1.0477$\pm$0.0506 & 1.0572$\pm$0.0527 & 0.9604$\pm$0.0022 & 0.9772$\pm$0.0012 & \\
        DiCNN \cite{he2019pansharpening} & 1.0525$\pm$0.0376 & 1.0812$\pm$0.0561 & 0.9594$\pm$0.0024 & 0.9781$\pm$0.0012 & \\
        MSDCNN \cite{wei2017multi} & 1.0472$\pm$0.0494 & 1.0413$\pm$0.0516 & 0.9612$\pm$0.0024 & 0.9782$\pm$0.0011 & \\
        FusionNet \cite{deng2020detail} & 0.9735$\pm$0.0473 & 0.9878$\pm$0.0497 & 0.9641$\pm$0.0021 & 0.9806$\pm$0.0011 & \\
        CTINN \cite{zhou2022pan} & 0.8251$\pm$0.0310 & 0.9595$\pm$0.0239 & 0.9772$\pm$0.0026 & 0.9803$\pm$0.0003 & \\
        LAGConv \cite{jin2022lagconv} & 0.7859$\pm$0.0330 & 0.6869$\pm$0.0252 & 0.9804$\pm$0.0009 & 0.9843$\pm$0.0005 & \\
        MMNet \cite{zhou2023memory} & 0.9292$\pm$0.0322 & 0.8117$\pm$0.0265 & 0.9690$\pm$0.0046 & 0.9859$\pm$0.0005 & \\
        DCFNet \cite{wu2021dynamic} & 0.8896$\pm$0.0353 & 0.8061$\pm$0.0306 & 0.9727$\pm$0.0022 & 0.9853$\pm$0.0005 & \\
        PanMamba \cite{he2025pan} & \underline{0.7433$\pm$0.0239} & \underline{0.6840$\pm$0.0245} & \underline{0.9824$\pm$0.0021} & \underline{0.9896$\pm$0.0004} & \\
        PanDiff \cite{meng2023pandiff} & 0.8881$\pm$0.0268 & 0.7461$\pm$0.0242 & 0.9727$\pm$0.0022 & 0.9887$\pm$0.0004 & \\
        PLRDiff \cite{rui2024unsupervised} & 2.6755$\pm$0.4609 & 2.7434$\pm$0.5485 & 0.7938$\pm$0.0367 & 0.9055$\pm$0.0170 & \\
        KSDiff (ours) & \textbf{0.6675$\pm$0.0299} & \textbf{0.5973$\pm$0.0229} & \textbf{0.9855$\pm$0.0020} & \textbf{0.9900$\pm$0.0005} & \\
        \bottomrule
        \end{tabular}
        \end{adjustbox}
        \label{gf2}
    \end{minipage}
    \hfill
    \begin{minipage}[t]{0.48\textwidth}
        \centering
        \caption{Result on the QB reduced-resolution dataset. The best results are highlighted in bold and the second best results are underlined.}
        \begin{adjustbox}{max width=\linewidth}
        \begin{tabular}{l|ccccccccc}
        \toprule
        \multirow{2}{*}{Method} & \multicolumn{4}{c}{\textbf{QuickBird}} \\
        & SAM $\downarrow$ ($\pm$ std) & ERGAS $\downarrow$ ($\pm$ std) & Q2\textsuperscript{n} $\uparrow$ ($\pm$ std) & SCC $\uparrow$ ($\pm$ std) \\
        \midrule
        BDSD-PC \cite{vivone2019robust}  & 8.2620$\pm$0.4583 & 7.5420$\pm$0.1819 & 0.8323$\pm$0.0226 & 0.9030$\pm$0.0040 \\
        MTF-GLP-FS \cite{vivone2018full}  & 8.1131$\pm$0.4371 & 7.5102$\pm$0.1772 & 0.8296$\pm$0.0202 & 0.8998$\pm$0.0044 \\
        BT-H \cite{aiazzi2006mtf}  & 7.1943$\pm$0.3470 & 7.4008$\pm$0.1873 & 0.8326$\pm$0.0197 & 0.9156$\pm$0.0034 \\
        PNN \cite{masi2016pansharpening}  & 5.2054$\pm$0.2152 & 4.4722$\pm$0.0835 & 0.9180$\pm$0.0210 & 0.9711$\pm$0.0027 \\
        DiCNN \cite{he2019pansharpening}  & 5.3795$\pm$0.2295 & 5.1354$\pm$0.1090 & 0.9042$\pm$0.0211 & 0.9621$\pm$0.0030 \\
        MSDCNN \cite{wei2017multi}  & 5.1471$\pm$0.2088 & 4.3828$\pm$0.0760 & 0.9180$\pm$0.0216 & 0.9689$\pm$0.0027 \\
        FusionNet \cite{deng2020detail}  & 4.9226$\pm$0.2029 & 4.1594$\pm$0.0718 & 0.9252$\pm$0.0202 & 0.9755$\pm$0.0023 \\
        CTINN \cite{zhou2022pan}  & 4.6583$\pm$0.1733 & 3.6969$\pm$0.0646 & 0.9320$\pm$0.0016 & 0.9829$\pm$0.0016 \\
        LAGConv \cite{jin2022lagconv}  & 4.5473$\pm$0.1855 & 3.8259$\pm$0.0938 & 0.9335$\pm$0.0196 & 0.9807$\pm$0.0020 \\
        MMNet \cite{zhou2023memory}  & 4.5568$\pm$0.1629 & \underline{3.6669$\pm$0.0679} & 0.9337$\pm$0.0210 & \underline{0.9829$\pm$0.0016} \\
        DCFNet \cite{wu2021dynamic}  & \underline{4.5383$\pm$0.1654} & 3.8315$\pm$0.0652 & 0.9325$\pm$0.0202 & 0.9741$\pm$0.0022 \\
        PanMamba \cite{he2025pan} & 4.6253$\pm$0.1773 & 4.2771$\pm$0.0708 & 0.9292$\pm$0.0224 & 0.9801$\pm$0.0022 \\
        PanDiff \cite{meng2023pandiff}  & 4.5754$\pm$0.1645 & 3.7422$\pm$0.0693 & \underline{0.9345$\pm$0.0202} & 0.9818$\pm$0.0202 \\
        PLRDiff \cite{rui2024unsupervised} & 26.8524$\pm$1.1153 & 47.4010$\pm$6.8066 & 0.2971$\pm$0.0550 & 0.6780$\pm$0.0336 \\
        KSDiff (ours) & \textbf{4.4747$\pm$0.1558} & \textbf{3.6289$\pm$0.0640} & \textbf{0.9365$\pm$0.0180} & \textbf{0.9839$\pm$0.0016} \\
        \bottomrule
        \end{tabular}
        \end{adjustbox}
    \label{qb}
    \end{minipage}
\end{table}

\begin{table}[t]
    \centering
    \small 
    \caption{Ablation study on the WV3 reduced-resolution dataset. The best results are highlighted.}
    \resizebox{\columnwidth}{!}{ 
    \begin{tabular}{l|cccc|c}
    \toprule
    Method & SAM $\downarrow$ ($\pm$ std) & ERGAS $\downarrow$ ($\pm$ std) & Q2\textsuperscript{n} $\uparrow$ ($\pm$ std) & SCC $\uparrow$ ($\pm$ std) & Runtime (s)\\
    \midrule
    Baseline Network & 3.1428$\pm$0.1333 & 2.2961$\pm$0.1349 & 0.9070$\pm$0.0191 & 0.9827$\pm$0.0014 & \textbf{0.035}\\
    w/o PLFE & 3.0071$\pm$0.1306 & 2.2367$\pm$0.1323 & 0.9119$\pm$0.0188 & 0.9838$\pm$0.0014 & 0.079\\
    w/o Structure-Aware & \multicolumn{4}{c|}{\textbf{Cannot Converge}} & -\\
    Separate-Training  & 2.9799$\pm$0.1295 & 2.1775$\pm$0.1266 & 0.9118$\pm$0.0189 & 0.9854$\pm$0.0012 & 0.077\\
    KSDiff (ours) & \textbf{2.8102$\pm$0.1147} & \textbf{2.0756$\pm$0.0973} & \textbf{0.9221$\pm$0.0183} & \textbf{0.9870$\pm$0.0010} & 0.077\\
    \bottomrule
    \end{tabular}}
    \label{ablation}
\end{table}

\begin{table}[htbp]
    \centering
    \small
    \begin{minipage}[t]{0.48\textwidth}
        \centering
        \caption{Result of replacing convolution modules in other backbones with KSDiff on WV3 reduced-resolution dataset.}
        \begin{adjustbox}{max width=\linewidth}
        \begin{tabular}{l|cccccccccc}
        \toprule
        \multirow{2}{*}{Method} & \multicolumn{4}{c}{\textbf{Metrics}} \\
        & SAM $\downarrow$ ($\pm$ std) & ERGAS $\downarrow$ ($\pm$ std) & Q2\textsuperscript{n} $\uparrow$ ($\pm$ std) & SCC $\uparrow$ ($\pm$ std) \\
        \midrule
        DiCNN \cite{he2019pansharpening}  & 3.5929$\pm$0.1705 & 2.6733$\pm$0.1482 & 0.9004$\pm$0.0195 & 0.9763$\pm$0.0016 & \\
        DiCNN+KSDiff  & 3.4154$\pm$0.1598 & 2.5278$\pm$0.1410 & 0.9052$\pm$0.0192 & 0.9791$\pm$0.0016 & \\
        \midrule
        FusionNet \cite{deng2020detail} & 3.3252$\pm$0.1560 & 2.4666$\pm$0.1441 & 0.9044$\pm$0.0202 & 0.9807$\pm$0.0015 &\\
        FusionNet+KSDiff  & 3.0622$\pm$0.1199 & 2.2725$\pm$0.1117 & 0.9111$\pm$0.0188 & 0.9837$\pm$0.0011 \\
        \midrule
        LAGNet \cite{jin2022lagconv} & 3.1042$\pm$0.1249 & 2.2999$\pm$0.1368 & 0.9098$\pm$0.0188 & 0.9830$\pm$0.0015 & \\
        LAGNet+KSDiff  & 2.9903$\pm$0.1277 & 2.1538$\pm$0.1352 & 0.9124$\pm$0.0193 & 0.9851$\pm$0.0014 & \\
        \bottomrule
        \end{tabular}
        \end{adjustbox}
        \label{replace}
    \end{minipage}
    \hfill
    \begin{minipage}[t]{0.48\textwidth}
        \centering
        \caption{Impact of low-rank core tensor size $(r_1, r_2, r_3, r_4)$ on the WV3 reduced-resolution dataset on FusionNet \cite{deng2020detail}.}
        \begin{adjustbox}{max width=\linewidth}
        \begin{tabular}{l|ccccccccc}
        \toprule
        \multirow{2}{*}{Size} & \multicolumn{4}{c}{\textbf{Metrics}} \\
        & SAM $\downarrow$ ($\pm$ std) & ERGAS $\downarrow$ ($\pm$ std) & Q2\textsuperscript{n} $\uparrow$ ($\pm$ std) & SCC $\uparrow$ ($\pm$ std) \\
        \midrule
        $(4, 4, 1, 1)$  & 3.0921$\pm$0.1202 & 2.2896$\pm$0.1119 & 0.9104$\pm$0.0188 & 0.9833$\pm$0.0012 \\
        $(4, 4, 2, 2)$ & 3.0622$\pm$0.1199 & 2.2725$\pm$0.1117 & 0.9111$\pm$0.0188 & 0.9837$\pm$0.0011 \\
        \midrule
        $(8, 8, 1, 1)$   & 3.1275$\pm$0.1256 & 2.3009$\pm$0.1174 & 0.9057$\pm$0.0200 & 0.9828$\pm$0.0011 \\
        $(8, 8, 2, 2)$  & 3.1040$\pm$0.1215 & 2.2988$\pm$0.1153 & 0.9096$\pm$0.0191 & 0.9833$\pm$0.0013 \\
        \midrule
        $(16, 16, 1, 1)$ & 3.1520$\pm$0.1262 & 2.3146$\pm$0.1180 & 0.9052$\pm$0.0195 & 0.9827$\pm$0.0015 \\
        $(16, 16, 2, 2)$  & 3.1475$\pm$0.1258 & 2.3106$\pm$0.1203 & 0.9055$\pm$0.0194 & 0.9829$\pm$0.0012 \\
        \bottomrule
        \end{tabular}
        \end{adjustbox}
        \label{core}
    \end{minipage}
\end{table}
\section{Conclusion}
In conclusion, we proposed KSDiff, a novel kernel space diffusion model that improves 
pansharpening by generating globally aware convolutional kernels through a diffusion process 
in latent space. By integrating a convolution core generator, a unified factor generator, 
and a structure-aware multi-head attention mechanism, KSDiff effectively balances high-quality 
fusion with fast inference. Extensive experiments on various datasets 
demonstrate its superior performance, offering a practical and scalable solution for remote sensing image fusion.

\newpage

{
\small
\bibliographystyle{ieeenat_fullname}
\bibliography{main}
}

\input{sec/X_suppl}

\end{document}

%% file: sec/X_suppl.tex
\clearpage
\setcounter{page}{1}
\maketitlesupplementary
\begin{abstract}
In this supplementary material, we first provide a more detailed explanation of specific components of our proposed method that could not be thoroughly described in the main paper, including the network we use in our main experiments and further information on latent encoders. We then elaborate on the experimental settings and implementation details. Finally, we present additional experimental results, including generalization assessment on WorldView-2 dataset, further visualizations on the WorldView-3, GaoFen-2, and QuickBird datasets, as well as ablation studies on more factors.
\end{abstract}

\section{Methods Explanation}
\label{sec: further methods}
\subsection{Pansharpening Network}
\begin{figure}[htbp]
    \centering
    \includegraphics[width=1.0\linewidth]{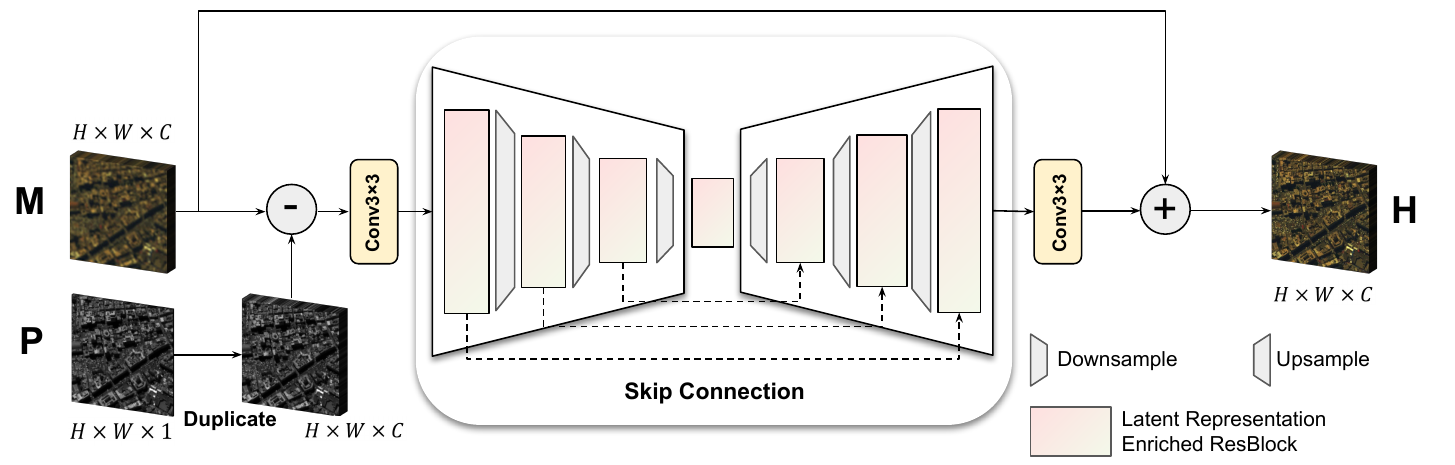}
    \caption{Baseline network.}
    \label{fig:backbone}
\end{figure}

The baseline network we build is demonstrated in \cref{fig:backbone}. The panchromatic (PAN) image $\mathbf{P} \in \mathbb{R}^{H \times W \times 1}$ is first duplicated along the channel dimension to match the number of channels in the low-resolution multispectral (LRMS) image $\mathbf{M} \in \mathbb{R}^{H \times W \times C}$. The duplicated PAN image is then subtracted by the LRMS image, and the resulting difference is fed into the network. 
After processing, the network output is added to the input LRMS image to produce the reconstructed high-resolution multispectral image $\mathbf{H} \in \mathbb{R}^{H \times W \times C}$, denoted as $\mathbf{H}_1$ or $\mathbf{H}_2$ as defined in the main text. To reduce spatial resolution and increase channel depth, downsampling layers are inserted between successive encoder blocks in the network. Conversely, in the decoder, upsampling operations with a scale factor of 2 are applied between adjacent layers to increase spatial resolution and reduce the number of channels. The skip connections adopt a concatenation-based strategy, similar to that of the standard U-Net \cite{ronneberger2015u}. Each block in the backbone corresponds to a ResBlock \cite{he2016deep}. Within each block, only the $3\times3$ convolution layer is reparameterized using the proposed KSDiff method. Given the multi-scale nature of the network, the latent representation is appropriately rescaled before being injected into each convolution layer. Specifically, shallower layers, which operate on higher spatial resolutions, use latent representations with larger scales, whereas deeper layers employ downsampled versions of the generated latent representation. Detailed parameter settings are provided in \cref{sec: exp details}.

\subsection{PLFE$_{2}$}
\begin{figure}[htbp]
    \centering
    \includegraphics[width=1.0\linewidth]{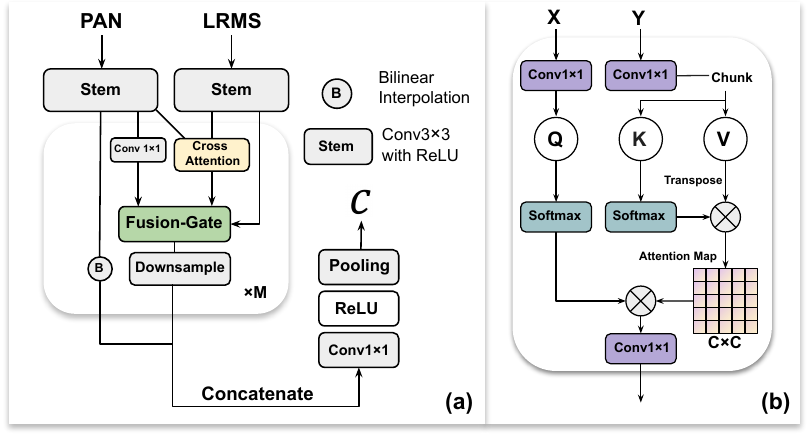}
    \caption{(a) $\mathrm{PLFE_2}$. (b) The cross attention in latent encoders.}
    \label{fig:plfe2_ca}
\end{figure}

As shown in \cref{fig:plfe2_ca} (a), the structure of $\mathrm{PLFE}_2$ is a simplified, halved version of $\mathrm{PLFE}_1$, which is presented in \cref{fig:plfe1} in the main text. All core components, including the cross-attention mechanism and the Fusion-Gate module, remain identical to those in $\mathrm{PLFE}_1$. The role of $\mathrm{PLFE}_2$ differs from $\mathrm{PLFE}_1$.
In $\mathrm{PLFE}_1$, the module encodes the ground-truth , PAN, LRMS images to obtain a latent representation $\mathbf{z}_0 \in \mathbb{R}^{N \times C_z}$. In contrast, $\mathrm{PLFE}_2$ produces a conditioning vector $\mathbf{c} \in \mathbb{R}^{N \times C_z}$ only with PAN and LRMS images as inputs, which is injected into the denoising network to guide the reverse diffusion process for latent reconstruction. The resulting latent representation is denoted as $\hat{\mathbf{z}}_0 \in \mathbb{R}^{N \times C_z}$, which is then fed into the kernel generator.

\begin{table*}[htbp]
    \centering
    \small  
    \renewcommand{\arraystretch}{1.6}  
    \caption{Summary of compared pansharpening methods.}
    \begin{tabularx}{\textwidth}{l|c|c|X}
        \toprule
        \toprule
        \textbf{Method} & \textbf{Category} & \textbf{Year} & \textbf{Introduction} \\
        \midrule
        BDSD-PC \cite{vivone2019robust} & CS & 2019 & Discusses the constraints of the band-dependent spatial-detail (BDSD) approach in images with over four spectral bands. \\
        MTF-GLP-FS \cite{vivone2018full} & MRA & 2018 & An iterative algorithm for estimating injection coefficients at full resolution in regression-based pansharpening. \\
        BT-H \cite{aiazzi2006mtf} & MRA & 2006 & A multiresolution pansharpening framework using a generalized Laplacian pyramid (GLP) with an MTF-adjusted filter to enhance spatial details while preserving spectral information. \\
        PNN \cite{masi2016pansharpening} & ML & 2016 & The first deep CNN architectures designed specifically for pansharpening. \\
        DiCNN \cite{he2019pansharpening} & ML & 2019 & Introduces a detail-injection mechanism to enhance spatial detail learning in CNNs. \\
        MSDCNN \cite{wei2017multi} & ML & 2017 & Employs multiscale and densely connected convolutional layers to extract both spatial and spectral features. \\
        FusionNet \cite{deng2020detail} & ML & 2020 & A residual learning-based network that emphasizes detail preservation for fusion tasks. \\
        CTINN \cite{zhou2022pan} & ML & 2022 & Use a customized transformer architecture and an information-lossless invertible neural module to model long-range dependencies. \\
        LAGConv \cite{jin2022lagconv} & ML & 2022 & Employ local-context adaptive convolution kernels and global harmonic bias to enhance image fusion. \\
        MMNet \cite{zhou2023memory} & ML & 2023 & A model-driven deep unfolding network with memory-augmentation. \\
        DCFNet \cite{wu2021dynamic} & ML & 2021 & A dynamic cross feature fusion network for pansharpening that integrates multi-scale branches and contextualized feature transfers. \\
        PanMamba \cite{he2025pan} & ML & 2025 & A state-space model based method for pansharpening. \\
        PanDiff \cite{meng2023pandiff} & Diffusion-based ML & 2023 & Employs a diffusion model to learn HRMS-IMS difference map distributions, guided by PAN/LRMS images with a modal intercalibration module. \\
        PLRDiff \cite{rui2024unsupervised} & Diffusion-based ML & 2024 & Proposes a low-rank diffusion model for pansharpening, combining a pre-trained diffusion model for image structure capture with a Bayesian approach for spectral preservation.\\
        \bottomrule
        \bottomrule
    \end{tabularx}
    \label{method_summary}
\end{table*}

\section{Details on Experiments}
\label{sec: exp details}
\subsection{Datasets}
We conducted experiments using datasets derived from WorldView-3 (WV3), QuickBird (QB), and GaoFen-2 (GF2) satellite imagery. 
These datasets consist of image patches cropped from full remote sensing scenes and are partitioned into training and testing subsets. The WV3 dataset contains four images from two geographic locations, Rio and Tripoli, captured in different seasons: Rio in January, Rio in May, Tripoli in March, and Tripoli in August. The right quarter of the Rio in May and Tripoli in August images was cropped to form the test set. The remaining areas of all images were used for training. This configuration avoids any spatial overlap and incorporates seasonal variation, thereby preventing both spatial and temporal leakage. For QB and GF2 datasets, each consists of a single image (QB Indianapolis and GF2 Guangzhou, respectively). The rightmost 1,000 pixels were cropped to form the test sets, while the rest of the image was used for training. Since the cropped test areas do not overlap with the training regions, spatial independence is maintained. Temporal leakage is not applicable in these cases, as only single-date imagery is available.

 The training data comprise simulated downsampling-based PAN/LRMS/GT triplets, with dimensions of $64 \times 64 \times 1$, $64 \times 64 \times C$, and $64 \times 64 \times C$, respectively. Specifically, the WV3 training set includes approximately 10{,}000 eight-channel samples ($C=8$), while the GF2 and QB training sets contain around 20{,}000 and 17{,}000 four-channel samples ($C=4$), respectively. For evaluation, the reduced-resolution test set for each satellite consists of 20 simulated PAN/LRMS/GT triplets representing diverse land cover types, with dimensions of $256 \times 256 \times 1$, $256 \times 256 \times C$, and $256 \times 256 \times C$, respectively. Additionally, a full-resolution test set is provided, comprising 20 original PAN/LRMS image pairs with spatial sizes of $512 \times 512$. All datasets and preprocessing procedures were obtained from the PanCollection repository~\cite{deng2022machine}.

 \subsection{Training Details}
\label{train_detail}

The proposed KSDiff framework is implemented using PyTorch 2.1.0 and Python 3.10. All experiments are conducted on a Linux system (Ubuntu 22.04) equipped with a single NVIDIA GeForce RTX 4090 GPU and CUDA version 12.1. We adopt the AdamW optimizer~\cite{loshchilov2017decoupled} with a learning rate of $0.8 \times 10^{-4}$ and a weight decay of $1 \times 10^{-4}$ for both the pre-training and diffusion training stages. The denoising diffusion process follows a cosine noise schedule~\cite{nichol2021improved}, without learning the variance as proposed in~\cite{nichol2021improved}. The total number of diffusion timesteps is set to 500 across all datasets. An exponential moving average (EMA) with a decay rate of 0.995 is employed. For the pre-training stage, the number of training iterations is set to 60k for all datasets (WV3, GF2, and QB). In the diffusion model training stage, the iteration counts are set to 50k, 70k, and 10k for WV3, GF2, and QB, respectively. The batch size during pre-training is fixed at 64 for all datasets. For diffusion model training, batch sizes are set to 32 for WV3, and 16 for both GF2 and QB. For DDIM sampling~\cite{song2020denoising}, we use 25 sampling steps uniformly across the three datasets.

The denoising network consists of five MLP blocks utilizing GELU activation~\cite{hendrycks2016gaussian}. The conditioning of the diffusion model was implemented via feature concatenation \cite{saharia2022image}. The latent encoders $\mathrm{PLFE}_1$ and $\mathrm{PLFE}_2$ are designed with three pyramid stages. The latent representation $\mathbf{z}$ has dimensions $N=16$ and $C_z=128$. The pansharpening network employs a U-Net \cite{ronneberger2015u} architecture with four encoder and four decoder blocks. The encoder begins with 32 channels, which are scaled by factors of 1, 2, 2, and 4 across successive blocks. The dimension of the latent representation $N$ is downsampled by the same factors before integration into kernel generation. For the kernel generator, the low-rank core tensor has dimensions $(r_1, r_2, r_3, r_4)$ set to $(4, 4, 2, 2)$ for the first three blocks and $(8, 8, 2, 2)$ for the fourth block.

\subsection{Compared Methods}
Table~\ref{method_summary} provides a brief overview of the pansharpening methods compared in the main text. We compare the proposed KSDiff with both traditional and recent machine learning (ML) approaches.
Three traditional methods are considered: BDSD-PC \cite{vivone2019robust}, MTF-GLP-FS \cite{vivone2018full}, and BT-H \cite{aiazzi2006mtf}. Nine machine learning methods without diffusion mechanisms are also included: PNN \cite{masi2016pansharpening}, DiCNN \cite{he2019pansharpening}, MSDCNN \cite{wei2017multi}, FusionNet \cite{deng2020detail}, CTINN \cite{zhou2022pan}, LAGConv \cite{jin2022lagconv}, MMNet \cite{zhou2023memory}, DCFNet \cite{wu2021dynamic}, and PanMamba \cite{he2025pan}.
Additionally, we include two recent diffusion-based ML methods: PanDiff \cite{meng2023pandiff} and PLRDiff \cite{rui2024unsupervised}. The runtimes of these methods (as reported in \cref{wv3} in the main text) are evaluated and compared using the WV3 reduced-resolution dataset.

\subsection{Ablation Study}
Continuing from \cref{ablation} in the main text, we provide the detailed implementation of our ablation studies. The experiment was implemented on the WV3 reduced-resolution dataset.
To assess the impact of the latent diffusion prior, we trained the baseline U-Net independently for 100{,}000 iterations using an $\ell_1$ loss function and a batch size of 64. For the ablation of our proposed PLFE module, we replaced the latent encoder with the one adopted in DiffIR~\cite{xia2023diffir}. In this setup, the PAN, LRMS, and GT images (for the pre-training stage), or the PAN and LRMS images (for the diffusion model training stage), were directly concatenated at the input of the encoder. The alternative encoder comprises six residual blocks.
All ablation experiments were trained until full convergence. Other training configurations, including the optimizer, learning rate, and weight decay, were kept consistent with those used in the main experiments, as illustrated in \cref{train_detail}.

\subsection{Other Backbones}
This section provides the implementation details of the experiments discussed in \cref{discuss} in the main text, where we investigate the effect of replacing standard convolutional operations with our proposed KSDiff kernel generation pipeline in other backbone networks. The experiment was implemented on the WV3 reduced-resolution dataset.
For the DiCNN model~\cite{he2019pansharpening}, which comprises a three-layer convolutional neural network with an inner channel dimension of 64, we modulated the middle convolution layer using the KSDiff mechanism. In the case of FusionNet~\cite{deng2020detail}, which contains four standard ResBlocks~\cite{he2016deep} with 32 inner channels each, we replaced the standard convolutional layers within the ResBlocks with their KSDiff-enhanced counterparts. For LAGNet~\cite{jin2022lagconv}, which already incorporates adaptive convolution kernels, we substituted the kernels in its adaptive convolution modules with diffusion-prior-enhanced versions generated by KSDiff.
For all three models, the batch size was consistently set to 64 for both the pre-training and diffusion model training stages. The size of the latent representation $\mathbf{z}$ is $N=16, C_z=32$. The low-rank core tensor size $(r_1, r_2, r_3, r_4)$ was fixed at $(4, 4, 2, 2)$, and the learning rate was set to $1 \times 10^{-4}$. All experiments were trained until full convergence.

\subsection{Core Tensor Size}
This section provides the detailed configuration of the experiments investigating the impact of low-rank core tensor sizes, as discussed in \cref{discuss} in the main text. The experiment was implemented on the WV3 reduced-resolution dataset. The inner channel dimension of FusionNet~\cite{deng2020detail} is fixed at 32, and all convolutional kernels have a size of $k=3$, making it a suitable and efficient setting for controlled comparative experiments.
All configurations presented in the main text were trained for 100{,}000 iterations in both the pre-training stage and the diffusion model training stage, with a batch size of 64 and a learning rate of $1\times10^{-4}$. The experimental results highlight the advantageous properties of leveraging low-rank representations and the high-dimensional structure of 4D tensors. We leave the exploration of these properties across a broader range of tasks, architectures, and parameter settings as a direction for future work.

\begin{table}[htbp]
    \centering
    \small
    \begin{minipage}[t]{0.48\textwidth}
        \centering
        \caption{Result on the GF2 full-resolution dataset. The best results are highlighted in bold and the second best results are underlined.}
        \begin{adjustbox}{max width=\linewidth}
        \begin{tabular}{l|cccccccccc}
        \toprule
        \multirow{2}{*}{Method} & \multicolumn{3}{c}{\textbf{GaoFen-2}} \\
        & $D_\lambda$ ($\pm$ std) & $D_s$ ($\pm$ std) & HQNR ($\pm$ std) &\\
        \midrule
        BDSD-PC \cite{vivone2019robust} & 0.0759$\pm$0.0067 & 0.1548$\pm$0.0063 & 0.7812$\pm$0.0091 \\
        MTF-GLP-FS \cite{vivone2018full} & 0.0336$\pm$0.0029 & 0.1404$\pm$0.0062 & 0.8309$\pm$0.0075 \\
        BT-H \cite{aiazzi2006mtf} & 0.0602$\pm$0.0043 & 0.1313$\pm$0.0043 & 0.8165$\pm$0.0068 \\
        PNN \cite{masi2016pansharpening} & 0.0367$\pm$0.0065 & 0.0943$\pm$0.0050 & 0.8726$\pm$0.0083 \\
        DiCNN \cite{he2019pansharpening} & 0.0413$\pm$0.0029 & 0.0992$\pm$0.0029 & 0.8636$\pm$0.0037 \\
        MSDCNN \cite{wei2017multi} & 0.0269$\pm$0.0029 & 0.0730$\pm$0.0021 & 0.9020$\pm$0.0029 \\
        FusionNet \cite{deng2020detail} & 0.0400$\pm$0.0028 & 0.1013$\pm$0.0030 & 0.8628$\pm$0.0041 \\
        CTINN \cite{zhou2022pan} & 0.0586$\pm$0.0058 & 0.1996$\pm$0.0033 & 0.8381$\pm$0.0053 \\
        LAGConv \cite{jin2022lagconv} & 0.0324$\pm$0.0029 & 0.0792$\pm$0.0030 & 0.8910$\pm$0.0046 \\
        MMNet \cite{zhou2023memory} & 0.0428$\pm$0.0067 & 0.1033$\pm$0.0029 & 0.8583$\pm$0.0060 \\
        DCFNet \cite{wu2021dynamic} & 0.0234$\pm$0.0026 & 0.0659$\pm$0.0021 & 0.9122$\pm$0.0027 \\
        PanMamba \cite{he2025pan} & \textbf{0.0231$\pm$0.0025} & \textbf{0.0572$\pm$0.0023} & \underline{0.9210$\pm$0.0028} \\
        PanDiff \cite{meng2023pandiff} & 0.0265$\pm$0.0044 & 0.0729$\pm$0.0023 & 0.9025$\pm$0.0047 \\
        PLRDiff \cite{rui2024unsupervised} & 0.2804$\pm$0.0292 & 0.1413$\pm$0.0057 & 0.6164$\pm$0.0234 \\
        KSDiff (ours) & \underline{0.0233$\pm$0.0028} & \underline{0.0588$\pm$0.0023} & \textbf{0.9257$\pm$0.0024} \\
        \midrule
        Ideal value & \textbf{0} & \textbf{0} & \textbf{1} \\
        \bottomrule
        \end{tabular}
        \end{adjustbox}
        \label{gf2_full_table}
    \end{minipage}
    \hfill
    \begin{minipage}[t]{0.48\textwidth}
        \centering
        \caption{Result on the QB full-resolution dataset. The best results are highlighted in bold and the second best results are underlined.}
        \begin{adjustbox}{max width=\linewidth}
        \begin{tabular}{l|ccccccccc}
        \toprule
        \multirow{2}{*}{Method} & \multicolumn{3}{c}{\textbf{QuickBird}} \\
        & $D_\lambda$ ($\pm$ std) & $D_s$ ($\pm$ std) & HQNR ($\pm$ std) &\\
        \midrule
        BDSD-PC \cite{vivone2019robust} & 0.1975$\pm$0.0075 & 0.1636$\pm$0.0108 & 0.6722$\pm$0.0129 \\
        MTF-GLP-FS \cite{vivone2018full} & 0.0489$\pm$0.0033 & 0.1383$\pm$0.0053 & 0.8199$\pm$0.0076 \\
        BT-H \cite{aiazzi2006mtf} & 0.2300$\pm$0.0161 & 0.1648$\pm$0.0037 & 0.6434$\pm$0.0144 \\
        PNN \cite{masi2016pansharpening} & 0.0569$\pm$0.0025 & 0.0624$\pm$0.0053 & 0.8844$\pm$0.0068 \\
        DiCNN \cite{he2019pansharpening} & 0.0920$\pm$0.0032 & 0.1067$\pm$0.0047 & 0.8114$\pm$0.0069 \\
        MSDCNN \cite{wei2017multi} & 0.0602$\pm$0.0034 & 0.0667$\pm$0.0065 & 0.8774$\pm$0.0087 \\
        FusionNet \cite{deng2020detail} & 0.0586$\pm$0.0042 & 0.0522$\pm$0.0020 & 0.8922$\pm$0.0049 \\
        CTINN \cite{zhou2022pan} & 0.1738$\pm$0.0074 & 0.0731$\pm$0.0053 & 0.7663$\pm$0.0097 \\
        LAGConv \cite{jin2022lagconv} & 0.0844$\pm$0.0053 & 0.0676$\pm$0.0030 & 0.8536$\pm$0.0040 \\
        MMNet \cite{zhou2023memory} & 0.0890$\pm$0.0114 & 0.0972$\pm$0.0085 & 0.8225$\pm$0.0071 \\
        DCFNet \cite{wu2021dynamic} & \underline{0.0454$\pm$0.0033} & 0.1239$\pm$0.0060 & 0.8360$\pm$0.0035 \\
        PanMamba \cite{he2025pan} & 0.0491$\pm$0.0028 & \underline{0.0443$\pm$0.0033} & \underline{0.9102$\pm$0.0029} \\
        PanDiff \cite{meng2023pandiff} & 0.0587$\pm$0.0039 & 0.0642$\pm$0.0056 & 0.8813$\pm$0.0093 \\
        PLRDiff \cite{rui2024unsupervised} & 0.7775$\pm$0.0233 & 0.3038$\pm$0.0204 & 0.1615$\pm$0.0200 \\
        KSDiff (ours) & \textbf{0.0380$\pm$0.0027} & \textbf{0.0426$\pm$0.0029} & \textbf{0.9130$\pm$0.0026} \\
        \midrule
        Ideal value & \textbf{0} & \textbf{0} & \textbf{1} \\
        \bottomrule
        \end{tabular}
        \end{adjustbox}
    \label{qb_full_table}
    \end{minipage}
\end{table}

\section{Additional Results}
\label{sec: add res}
\subsection{Main Results}
\begin{figure*}[htbp]
    \centering
    \includegraphics[width=1.0\linewidth]{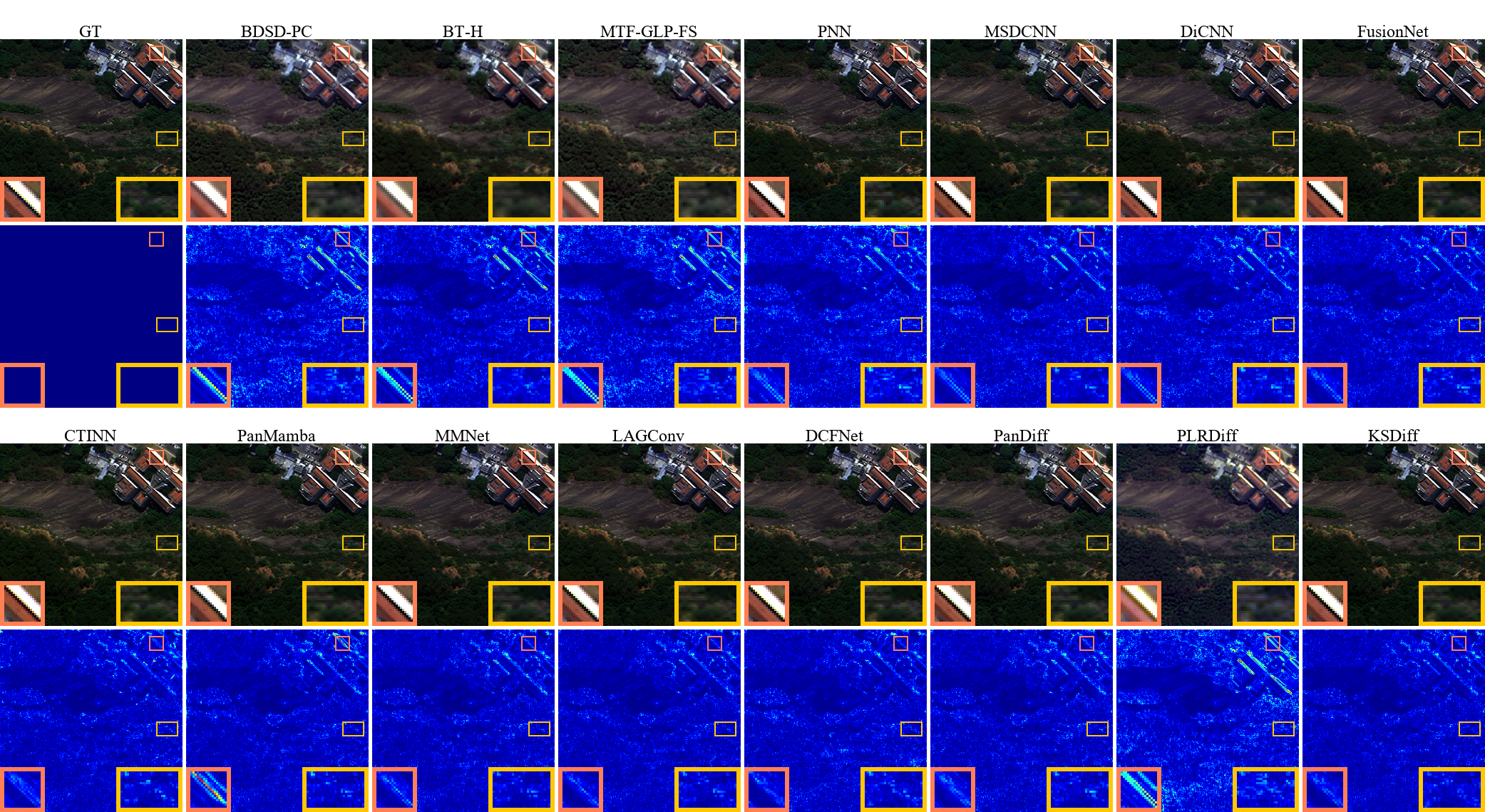}
    \caption{Comparison of qualitative results for representative methods on the WV3 reduced-resolution dataset.}
    \label{wv3_1}
\end{figure*}

\begin{figure*}[htbp]
    \centering
    \includegraphics[width=1.0\linewidth]{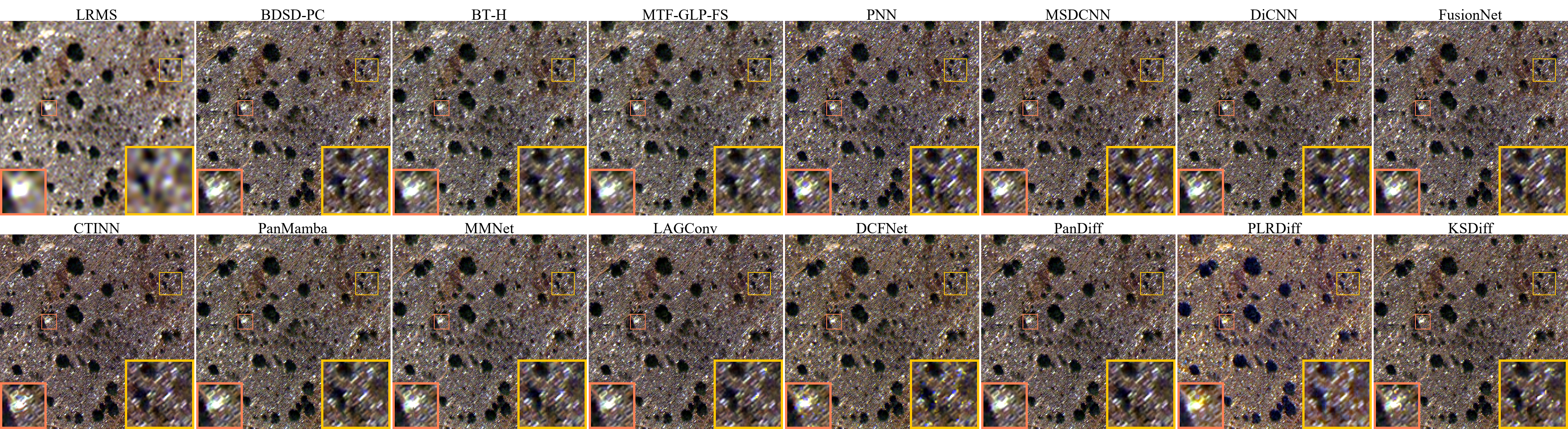}
    \caption{Comparison of qualitative results for representative methods on the WV3 full-resolution dataset.}
    \label{wv3_full}
\end{figure*}

\begin{figure*}[htbp]
    \centering
    \includegraphics[width=1.0\linewidth]{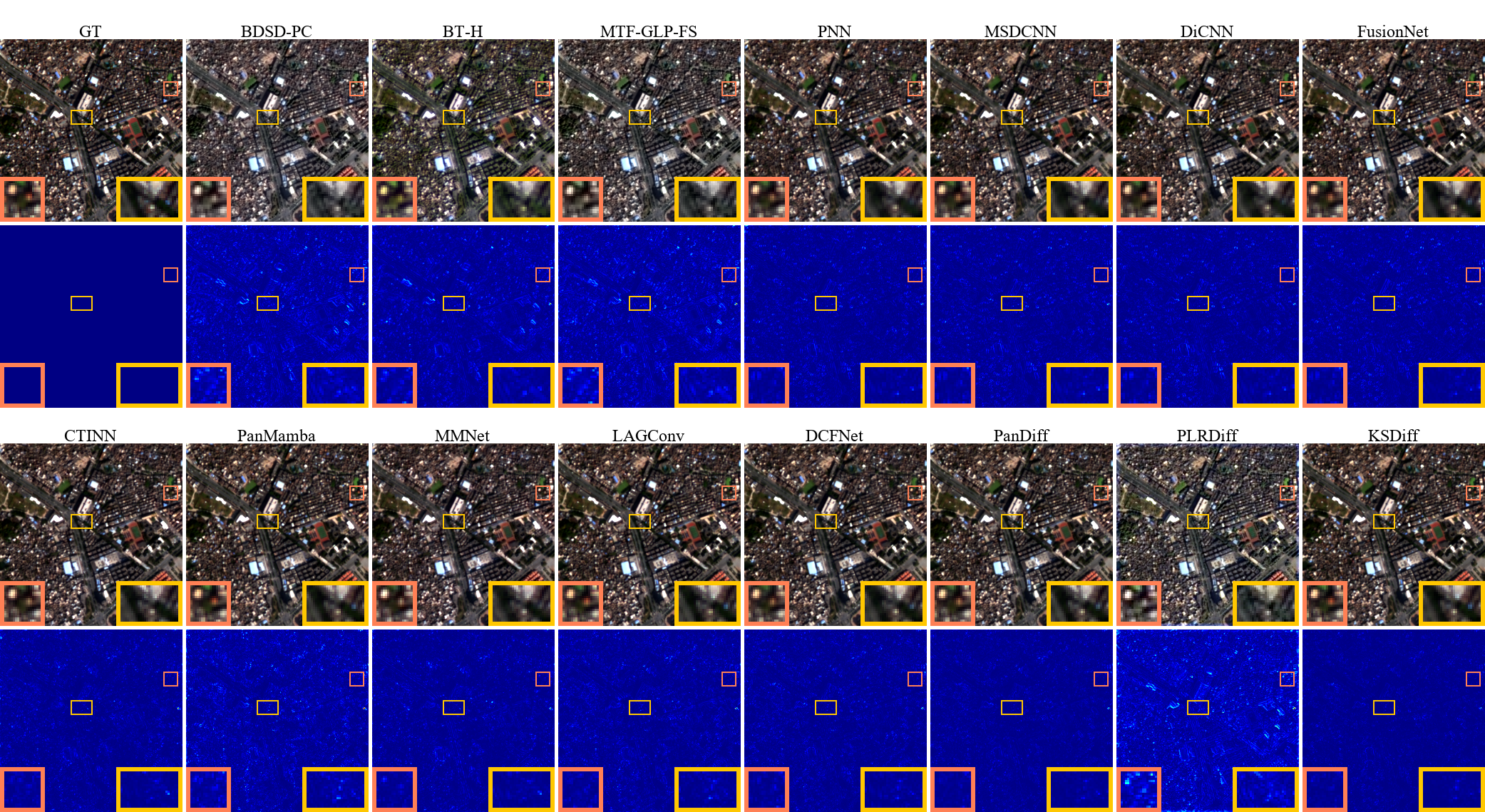}
    \caption{Comparison of qualitative results for representative methods on the GF2 reduced-resolution dataset.}
    \label{gf2_2}
\end{figure*}

\begin{figure*}[htbp]
    \centering
    \includegraphics[width=1.0\linewidth]{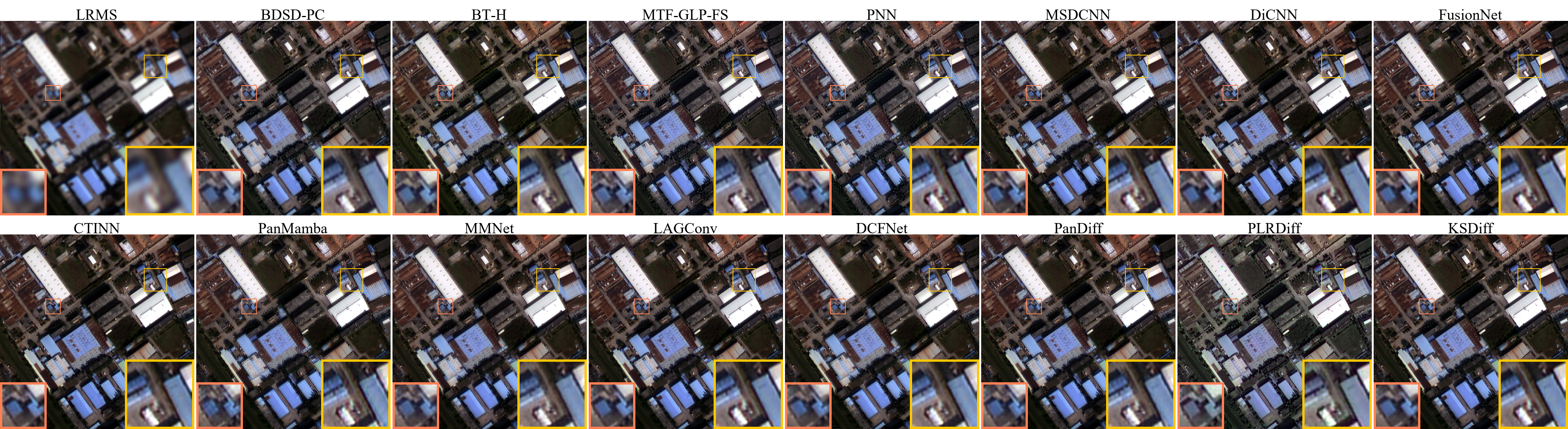}
    \caption{Comparison of qualitative results for representative methods on the GF2 full-resolution dataset.}
    \label{gf2_full}
\end{figure*}

\begin{figure*}[htbp]
    \centering
    \includegraphics[width=1.0\linewidth]{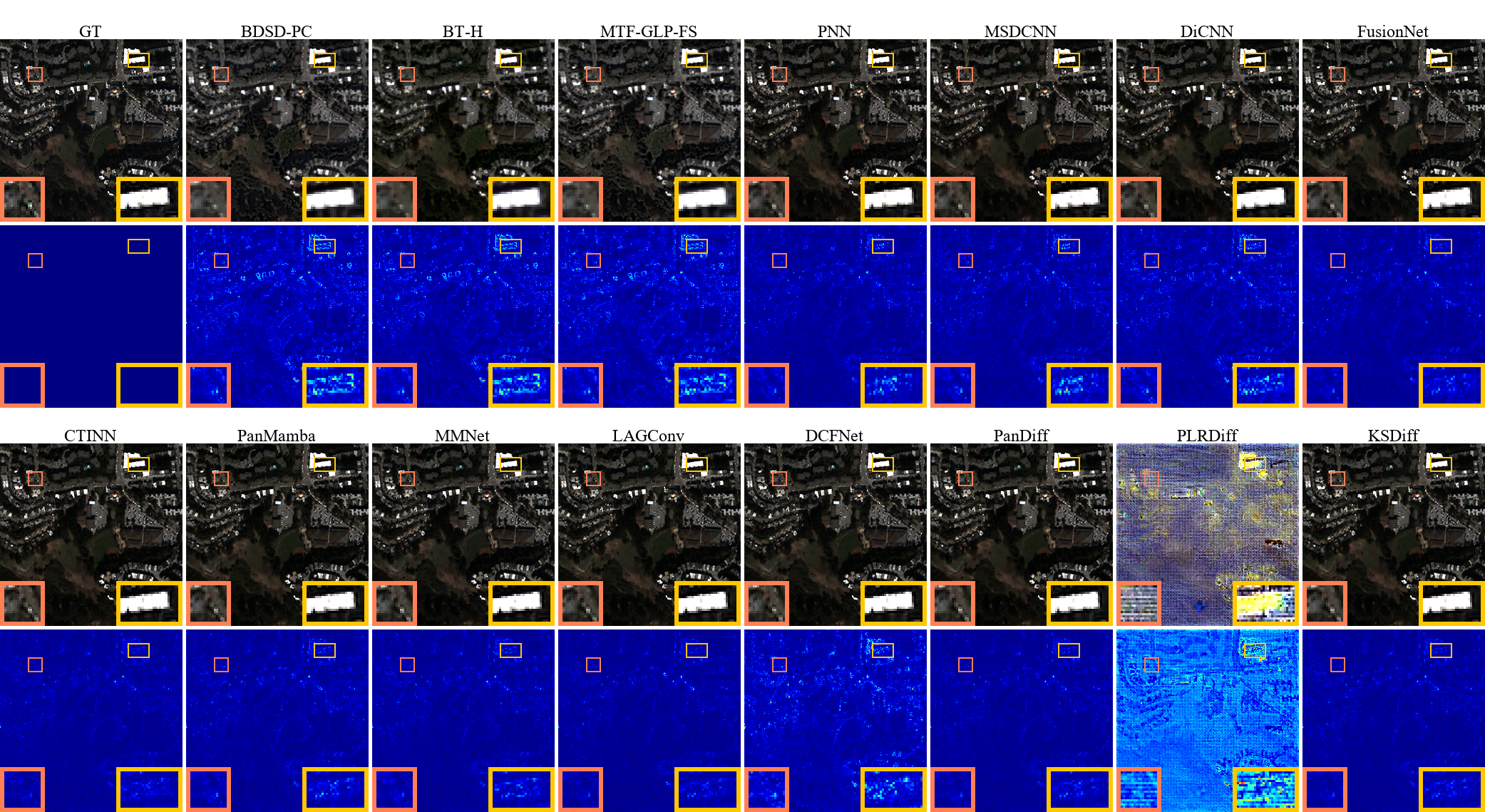}
    \caption{Comparison of qualitative results for representative methods on the QB reduced-resolution dataset.}
    \label{qb_1}
\end{figure*}

\begin{figure*}[htbp]
    \centering
    \includegraphics[width=1.0\linewidth]{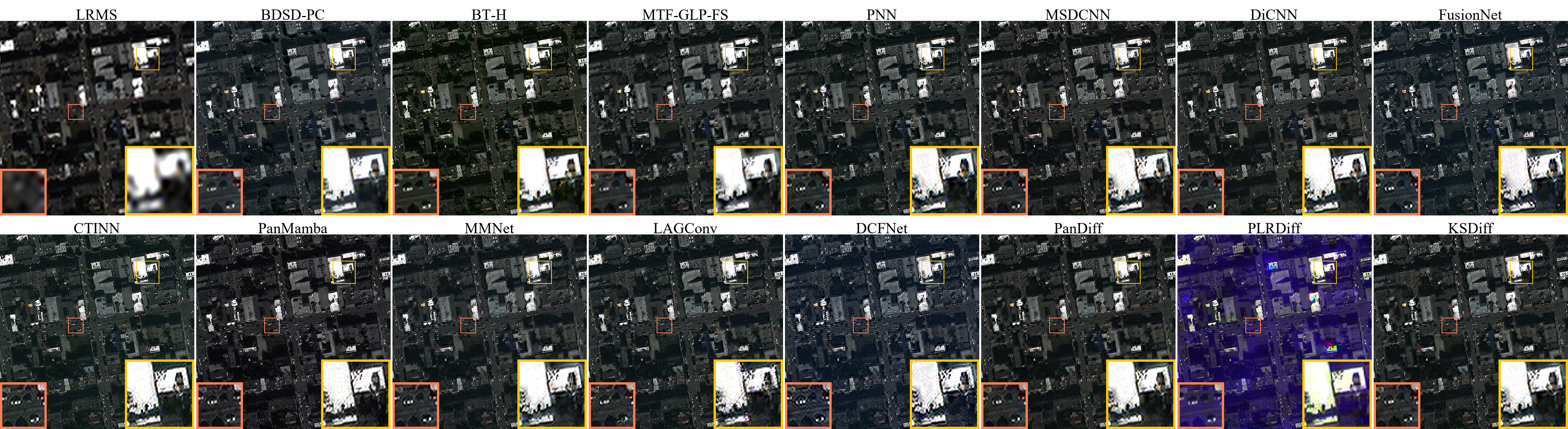}
    \caption{Comparison of qualitative results for representative methods on the QB full-resolution dataset.}
    \label{qb_full}
\end{figure*}

\cref{gf2_full_table} and \cref{qb_full_table} present the quantitative performance benchmarks on the full-resolution GF2 and QB datasets. The results indicate that the proposed KSDiff method exhibits strong generalization capabilities across different data domains.
\cref{wv3_1} to \cref{qb_full} provide qualitative comparisons of visual outputs generated by various methods on representative samples from the WV3, GF2, and QB datasets. For the reduced-resolution data, corresponding error maps between the predicted outputs and ground-truth references are also included. The visual and quantitative comparisons consistently demonstrate that KSDiff produces results that are highly consistent with the ground-truth. Furthermore, the method shows robust performance in scenarios lacking reference images, attributed to the diffusion model's powerful distribution estimation capability.

\begin{table}[htbp]
    \centering
    \small
    \begin{minipage}[t]{0.48\textwidth}
    \centering
    \caption{Generalization of DL-based methods on WV2 dataset.}
    \begin{adjustbox}{max width=\linewidth}
    \begin{tabular}{l|cccc}
        \toprule
        \multirow{2}{*}{Method} & \multicolumn{4}{c}{\textbf{WorldView-2}} \\
        & SAM ($\pm$ std) & ERGAS ($\pm$ std) & Q2\textsuperscript{n} ($\pm$ std) & SCC ($\pm$ std) \\
        \midrule
        PNN \cite{masi2016pansharpening}           & 7.1158$\pm$0.3758 & 5.6152$\pm$0.2108 & 0.7619$\pm$0.0207 & 0.8782$\pm$0.0039 \\
        DiCNN \cite{he2019pansharpening}        & 6.9216$\pm$0.1766 & 6.2507$\pm$0.1285 & 0.7205$\pm$0.0167 & 0.8552$\pm$0.0065 \\
        MSDCNN \cite{wei2017multi}       & 6.0064$\pm$0.1425 & 4.7438$\pm$0.1104 & 0.8241$\pm$0.0179 & 0.8972$\pm$0.0024 \\
        FusionNet \cite{deng2020detail}    & 6.4257$\pm$0.1923 & 5.1363$\pm$0.1152 & 0.7961$\pm$0.0165 & 0.8746$\pm$0.0030 \\
        CTINN  \cite{zhou2022pan}       & 6.4103$\pm$0.1331 & 4.6435$\pm$0.0847 & 0.8172$\pm$0.0195 & 0.9147$\pm$0.0023 \\
        LAGConv \cite{jin2022lagconv}      & 6.9545$\pm$0.1059 & 5.3262$\pm$0.0712 & 0.8054$\pm$0.0187 & 0.9125$\pm$0.0023 \\
        MMNet \cite{zhou2023memory}        & 6.6109$\pm$0.0717 & 5.2213$\pm$0.0477 & 0.8143$\pm$0.0177 & 0.9136$\pm$0.0045 \\
        DCFNet \cite{wu2021dynamic}       & \underline{5.6194$\pm$0.1350} & \underline{4.4887$\pm$0.0841} & \underline{0.8292$\pm$0.0182} & \underline{0.9154$\pm$0.0019} \\
        KSDiff (ours) & \textbf{5.1944$\pm$0.1197} & \textbf{4.1052$\pm$0.0796} & \textbf{0.8485$\pm$0.0185} & \textbf{0.9288$\pm$0.0017} \\
        \midrule
        Ideal value   & \textbf{0} & \textbf{0} & \textbf{1} & \textbf{1} \\
        \bottomrule
    \end{tabular}
    \end{adjustbox}
    \label{tab:wv2}
\end{minipage}
    \hfill
\begin{minipage}[t]{0.48\textwidth}
    \centering
    \caption{Ablation study on the trade-off hyperparameter $\lambda$ in the loss function of the diffusion model training stage.}
    \begin{adjustbox}{max width=\linewidth}
    \begin{tabular}{c|cccc}
        \toprule
        \multirow{2}{*}{$\lambda$} & \multicolumn{4}{c}{\textbf{WorldView-3}} \\
        & SAM↓ ($\pm$ std) & ERGAS↓ ($\pm$ std) & Qn↑ ($\pm$ std) & SCC↑ ($\pm$ std) \\
        \midrule
        0.1   & 2.8232$\pm$0.1167 & 2.0858$\pm$0.1021 & 0.9167$\pm$0.0186 & 0.9870$\pm$0.0010 \\
        0.2   & 2.8159$\pm$0.1155 & 2.0801$\pm$0.1006 & 0.9168$\pm$0.0188 & 0.9871$\pm$0.0009 \\
        1     & 2.8102$\pm$0.1147 & 2.0756$\pm$0.0973 & 0.9221$\pm$0.0183 & 0.9870$\pm$0.0010 \\
        5     & 2.8078$\pm$0.1147 & 2.0790$\pm$0.1015 & 0.9203$\pm$0.0185 & 0.9870$\pm$0.0010 \\
        10    & 2.8106$\pm$0.1149 & 2.0877$\pm$0.1016 & 0.9189$\pm$0.0190 & 0.9869$\pm$0.0010 \\
        \midrule
        Ideal value & \textbf{0} & \textbf{0} & \textbf{1} & \textbf{1} \\
        \bottomrule
    \end{tabular}
    \end{adjustbox}
    \label{tab:ablation_lambda}
\end{minipage}
\end{table}

\subsection{Generalization}
To assess the generalization capability of deep learning–based methods, we evaluated models trained on the WV3 dataset using 20 reduced resolutions from the WorldView-2 dataset. As shown in \cref{tab:wv2}, the quantitative results reveal that the KSDiff method consistently outperforms others across all four evaluation metrics, demonstrating the strong generalization ability of our approach.

\subsection{Effect of Hyperparameter $\lambda$}
For the total loss design in the diffusion model training stage,  
$\mathcal{L} = \mathcal{L}_{\text{diff}} + \lambda \mathcal{L}_{\text{reg}}$, 
we initially set \(\lambda = 1\), which was found to be empirically effective. 
Throughout our experiments, no issues such as gradient collapse were observed. 
To further examine the influence of \(\lambda\), we conducted an ablation study in which all models were trained until full convergence. 
The results, summarized in \cref{tab:ablation_lambda}, demonstrate that the overall performance remains relatively consistent across a range of $\lambda$ values, indicating that our method is not particularly sensitive to this hyperparameter.

%% file: main.bib
@String(CVPR= {IEEE Conf. Comput. Vis. Pattern Recog.})

@String(IJCAI = {IJCAI})

@String(AAAI = {AAAI})

@String(CVPR  = {CVPR})

@article{wu2023framelet,
  title={A framelet sparse reconstruction method for pansharpening with guaranteed convergence},
  author={Wu, Zhong-Cheng and Huang, Ting-Zhu and Deng, Liang-Jian and Vivone, Gemine},
  journal={Inverse Problems and Imaging},
  volume={17},
  number={6},
  pages={1277--1300},
  year={2023},
  publisher={Inverse Problems and Imaging}
}

@article{wu2023lrtcfpan,
  title={LRTCFPan: Low-rank tensor completion based framework for pansharpening},
  author={Wu, Zhong-Cheng and Huang, Ting-Zhu and Deng, Liang-Jian and Huang, Jie and Chanussot, Jocelyn and Vivone, Gemine},
  journal={IEEE Transactions on Image Processing},
  volume={32},
  pages={1640--1655},
  year={2023},
  publisher={IEEE}
}

@article{vivone2017regression,
  title={A regression-based high-pass modulation pansharpening approach},
  author={Vivone, Gemine and Restaino, Rocco and Chanussot, Jocelyn},
  journal={IEEE Transactions on geoscience and remote sensing},
  volume={56},
  number={2},
  pages={984--996},
  year={2017},
  publisher={IEEE}
}

@article{otazu2005introduction,
  title={Introduction of sensor spectral response into image fusion methods. Application to wavelet-based methods},
  author={Otazu, Xavier and Gonz{\'a}lez-Aud{\'\i}cana, Mar{\'\i}a and Fors, Octavi and N{\'u}{\~n}ez, Jorge},
  journal={IEEE Transactions on Geoscience and Remote Sensing},
  volume={43},
  number={10},
  pages={2376--2385},
  year={2005},
  publisher={IEEE}
}

@article{vivone2018full,
  title={Full scale regression-based injection coefficients for panchromatic sharpening},
  author={Vivone, Gemine and Restaino, Rocco and Chanussot, Jocelyn},
  journal={IEEE Transactions on Image Processing},
  volume={27},
  number={7},
  pages={3418--3431},
  year={2018},
  publisher={IEEE}
}

@article{meng2016pansharpening,
  title={Pansharpening with a guided filter based on three-layer decomposition},
  author={Meng, Xiangchao and Li, Jie and Shen, Huanfeng and Zhang, Liangpei and Zhang, Hongyan},
  journal={Sensors},
  volume={16},
  number={7},
  pages={1068},
  year={2016},
  publisher={MDPI}
}

@article{kwarteng1989extracting,
  title={Extracting spectral contrast in Landsat Thematic Mapper image data using selective principal component analysis},
  author={Kwarteng, P and Chavez, A},
  journal={Photogramm. Eng. Remote Sens},
  volume={55},
  number={1},
  pages={339--348},
  year={1989}
}

@article{wu2017post,
  title={A post-classification change detection method based on iterative slow feature analysis and Bayesian soft fusion},
  author={Wu, Chen and Du, Bo and Cui, Xiaohui and Zhang, Liangpei},
  journal={Remote Sensing of Environment},
  volume={199},
  pages={241--255},
  year={2017},
  publisher={Elsevier}
}

@article{yuan2021review,
  title={A review of deep learning methods for semantic segmentation of remote sensing imagery},
  author={Yuan, Xiaohui and Shi, Jianfang and Gu, Lichuan},
  journal={Expert Systems with Applications},
  volume={169},
  pages={114417},
  year={2021},
  publisher={Elsevier}
}

@article{masi2016pansharpening,
  title={Pansharpening by convolutional neural networks},
  author={Masi, Giuseppe and Cozzolino, Davide and Verdoliva, Luisa and Scarpa, Giuseppe},
  journal={Remote Sensing},
  volume={8},
  number={7},
  pages={594},
  year={2016},
  publisher={MDPI}
}

@inproceedings{yang2017pannet,
  title={PanNet: A deep network architecture for pan-sharpening},
  author={Yang, Junfeng and Fu, Xueyang and Hu, Yuwen and Huang, Yue and Ding, Xinghao and Paisley, John},
  booktitle={Proceedings of the IEEE international conference on computer vision},
  pages={5449--5457},
  year={2017}
}

@article{he2019pansharpening,
  title={Pansharpening via detail injection based convolutional neural networks},
  author={He, Lin and Rao, Yizhou and Li, Jun and Chanussot, Jocelyn and Plaza, Antonio and Zhu, Jiawei and Li, Bo},
  journal={IEEE Journal of Selected Topics in Applied Earth Observations and Remote Sensing},
  volume={12},
  number={4},
  pages={1188--1204},
  year={2019},
  publisher={IEEE}
}

@article{deng2020detail,
  title={Detail injection-based deep convolutional neural networks for pansharpening},
  author={Deng, Liang-Jian and Vivone, Gemine and Jin, Cheng and Chanussot, Jocelyn},
  journal={IEEE Transactions on Geoscience and Remote Sensing},
  volume={59},
  number={8},
  pages={6995--7010},
  year={2020},
  publisher={IEEE}
}

@article{wald1997fusion,
  title={Fusion of satellite images of different spatial resolutions: Assessing the quality of resulting images},
  author={Wald, Lucien and Ranchin, Thierry and Mangolini, Marc},
  journal={Photogrammetric engineering and remote sensing},
  volume={63},
  number={6},
  pages={691--699},
  year={1997}
}

@article{deng2022machine,
  title={Machine learning in pansharpening: A benchmark, from shallow to deep networks},
  author={Deng, Liang-Jian and Vivone, Gemine and Paoletti, Mercedes E and Scarpa, Giuseppe and He, Jiang and Zhang, Yongjun and Chanussot, Jocelyn and Plaza, Antonio},
  journal={IEEE Geoscience and Remote Sensing Magazine},
  volume={10},
  number={3},
  pages={279--315},
  year={2022},
  publisher={IEEE}
}

@inproceedings{boardman1993automating,
  title={Automating spectral unmixing of AVIRIS data using convex geometry concepts},
  author={Boardman, Joseph W},
  booktitle={JPL, Summaries of the 4th Annual JPL Airborne Geoscience Workshop. Volume 1: AVIRIS Workshop},
  year={1993}
}

@book{wald2002data,
  title={Data fusion: definitions and architectures: fusion of images of different spatial resolutions},
  author={Wald, Lucien},
  year={2002},
  publisher={Presses des MINES}
}

@article{garzelli2009hypercomplex,
  title={Hypercomplex quality assessment of multi/hyperspectral images},
  author={Garzelli, Andrea and Nencini, Filippo},
  journal={IEEE Geoscience and Remote Sensing Letters},
  volume={6},
  number={4},
  pages={662--665},
  year={2009},
  publisher={IEEE}
}

@article{zhou1998wavelet,
  title={A wavelet transform method to merge Landsat TM and SPOT panchromatic data},
  author={Zhou, Jie and Civco, Daniel L and Silander, John A},
  journal={International journal of remote sensing},
  volume={19},
  number={4},
  pages={743--757},
  year={1998},
  publisher={Taylor \& Francis}
}

@article{arienzo2022full,
  title={Full-resolution quality assessment of pansharpening: Theoretical and hands-on approaches},
  author={Arienzo, Alberto and Vivone, Gemine and Garzelli, Andrea and Alparone, Luciano and Chanussot, Jocelyn},
  journal={IEEE Geoscience and Remote Sensing Magazine},
  volume={10},
  number={3},
  pages={168--201},
  year={2022},
  publisher={IEEE}
}

@article{loshchilov2017decoupled,
  title={Decoupled weight decay regularization},
  author={Loshchilov, Ilya and Hutter, Frank},
  journal={arXiv preprint arXiv:1711.05101},
  year={2017}
}

@inproceedings{wu2021dynamic,
  title={Dynamic cross feature fusion for remote sensing pansharpening},
  author={Wu, Xiao and Huang, Ting-Zhu and Deng, Liang-Jian and Zhang, Tian-Jing},
  booktitle={Proceedings of the IEEE/CVF International Conference on Computer Vision},
  pages={14687--14696},
  year={2021}
}

@article{liang2022pmacnet,
  title={PMACNet: Parallel multiscale attention constraint network for pan-sharpening},
  author={Liang, Yixun and Zhang, Ping and Mei, Yang and Wang, Tingqi},
  journal={IEEE Geoscience and Remote Sensing Letters},
  volume={19},
  pages={1--5},
  year={2022},
  publisher={IEEE}
}

@inproceedings{jin2022lagconv,
  title={LAGConv: Local-context adaptive convolution kernels with global harmonic bias for pansharpening},
  author={Jin, Zi-Rong and Zhang, Tian-Jing and Jiang, Tai-Xiang and Vivone, Gemine and Deng, Liang-Jian},
  booktitle={Proceedings of the AAAI conference on artificial intelligence},
  volume={36},
  number={1},
  pages={1113--1121},
  year={2022}
}

@inproceedings{duan2024content,
  title={Content-adaptive non-local convolution for remote sensing pansharpening},
  author={Duan, Yule and Wu, Xiao and Deng, Haoyu and Deng, Liang-Jian},
  booktitle={Proceedings of the IEEE/CVF Conference on Computer Vision and Pattern Recognition},
  pages={27738--27747},
  year={2024}
}

@inproceedings{tan2024revisiting,
  title={Revisiting Spatial-Frequency Information Integration from a Hierarchical Perspective for Panchromatic and Multi-Spectral Image Fusion},
  author={Tan, Jiangtong and Huang, Jie and Zheng, Naishan and Zhou, Man and Yan, Keyu and Hong, Danfeng and Zhao, Feng},
  booktitle={Proceedings of the IEEE/CVF Conference on Computer Vision and Pattern Recognition},
  pages={25922--25931},
  year={2024}
}

@article{ho2020denoising,
  title={Denoising diffusion probabilistic models},
  author={Ho, Jonathan and Jain, Ajay and Abbeel, Pieter},
  journal={Advances in neural information processing systems},
  volume={33},
  pages={6840--6851},
  year={2020}
}

@article{song2020denoising,
  title={Denoising diffusion implicit models},
  author={Song, Jiaming and Meng, Chenlin and Ermon, Stefano},
  journal={arXiv preprint arXiv:2010.02502},
  year={2020}
}

@article{zheng2023dpm,
  title={Dpm-solver-v3: Improved diffusion ode solver with empirical model statistics},
  author={Zheng, Kaiwen and Lu, Cheng and Chen, Jianfei and Zhu, Jun},
  journal={Advances in Neural Information Processing Systems},
  volume={36},
  pages={55502--55542},
  year={2023}
}

@article{lu2022dpm,
  title={Dpm-solver: A fast ode solver for diffusion probabilistic model sampling in around 10 steps},
  author={Lu, Cheng and Zhou, Yuhao and Bao, Fan and Chen, Jianfei and Li, Chongxuan and Zhu, Jun},
  journal={Advances in Neural Information Processing Systems},
  volume={35},
  pages={5775--5787},
  year={2022}
}

@article{yue2023resshift,
  title={Resshift: Efficient diffusion model for image super-resolution by residual shifting},
  author={Yue, Zongsheng and Wang, Jianyi and Loy, Chen Change},
  journal={Advances in Neural Information Processing Systems},
  volume={36},
  pages={13294--13307},
  year={2023}
}

@misc{kingma2013auto,
  title={Auto-encoding variational bayes},
  author={Kingma, Diederik P and Welling, Max and others},
  year={2013},
  publisher={Banff, Canada}
}

@article{wang2024exploiting,
  title={Exploiting diffusion prior for real-world image super-resolution},
  author={Wang, Jianyi and Yue, Zongsheng and Zhou, Shangchen and Chan, Kelvin CK and Loy, Chen Change},
  journal={International Journal of Computer Vision},
  volume={132},
  number={12},
  pages={5929--5949},
  year={2024},
  publisher={Springer}
}

@inproceedings{xia2023diffir,
  title={Diffir: Efficient diffusion model for image restoration},
  author={Xia, Bin and Zhang, Yulun and Wang, Shiyin and Wang, Yitong and Wu, Xinglong and Tian, Yapeng and Yang, Wenming and Van Gool, Luc},
  booktitle={Proceedings of the IEEE/CVF International Conference on Computer Vision},
  pages={13095--13105},
  year={2023}
}

@article{chung2022diffusion,
  title={Diffusion posterior sampling for general noisy inverse problems},
  author={Chung, Hyungjin and Kim, Jeongsol and Mccann, Michael T and Klasky, Marc L and Ye, Jong Chul},
  journal={arXiv preprint arXiv:2209.14687},
  year={2022}
}

@article{wang2024neural,
  title={Neural network diffusion},
  author={Wang, Kai and Tang, Dongwen and Zeng, Boya and Yin, Yida and Xu, Zhaopan and Zhou, Yukun and Zang, Zelin and Darrell, Trevor and Liu, Zhuang and You, Yang},
  journal={arXiv preprint arXiv:2402.13144},
  year={2024}
}

@article{karras2022elucidating,
  title={Elucidating the design space of diffusion-based generative models},
  author={Karras, Tero and Aittala, Miika and Aila, Timo and Laine, Samuli},
  journal={Advances in neural information processing systems},
  volume={35},
  pages={26565--26577},
  year={2022}
}

@article{shaul2023bespoke,
  title={Bespoke solvers for generative flow models},
  author={Shaul, Neta and Perez, Juan and Chen, Ricky TQ and Thabet, Ali and Pumarola, Albert and Lipman, Yaron},
  journal={arXiv preprint arXiv:2310.19075},
  year={2023}
}

@article{liu2022flow,
  title={Flow straight and fast: Learning to generate and transfer data with rectified flow},
  author={Liu, Xingchao and Gong, Chengyue and Liu, Qiang},
  journal={arXiv preprint arXiv:2209.03003},
  year={2022}
}

@article{song2023consistency,
  title={Consistency models},
  author={Song, Yang and Dhariwal, Prafulla and Chen, Mark and Sutskever, Ilya},
  year={2023}
}

@article{song2020score,
  title={Score-based generative modeling through stochastic differential equations},
  author={Song, Yang and Sohl-Dickstein, Jascha and Kingma, Diederik P and Kumar, Abhishek and Ermon, Stefano and Poole, Ben},
  journal={arXiv preprint arXiv:2011.13456},
  year={2020}
}

@inproceedings{rombach2022high,
  title={High-resolution image synthesis with latent diffusion models},
  author={Rombach, Robin and Blattmann, Andreas and Lorenz, Dominik and Esser, Patrick and Ommer, Bj{\"o}rn},
  booktitle={Proceedings of the IEEE/CVF conference on computer vision and pattern recognition},
  pages={10684--10695},
  year={2022}
}

@inproceedings{esser2024scaling,
  title={Scaling rectified flow transformers for high-resolution image synthesis},
  author={Esser, Patrick and Kulal, Sumith and Blattmann, Andreas and Entezari, Rahim and M{\"u}ller, Jonas and Saini, Harry and Levi, Yam and Lorenz, Dominik and Sauer, Axel and Boesel, Frederic and others},
  booktitle={Forty-first international conference on machine learning},
  year={2024}
}

@article{albergo2023stochastic,
  title={Stochastic interpolants: A unifying framework for flows and diffusions},
  author={Albergo, Michael S and Boffi, Nicholas M and Vanden-Eijnden, Eric},
  journal={arXiv preprint arXiv:2303.08797},
  year={2023}
}

@article{lipman2022flow,
  title={Flow matching for generative modeling},
  author={Lipman, Yaron and Chen, Ricky TQ and Ben-Hamu, Heli and Nickel, Maximilian and Le, Matt},
  journal={arXiv preprint arXiv:2210.02747},
  year={2022}
}

@article{zhou2023denoising,
  title={Denoising diffusion bridge models},
  author={Zhou, Linqi and Lou, Aaron and Khanna, Samar and Ermon, Stefano},
  journal={arXiv preprint arXiv:2309.16948},
  year={2023}
}

@article{de2021diffusion,
  title={Diffusion schr{\"o}dinger bridge with applications to score-based generative modeling},
  author={De Bortoli, Valentin and Thornton, James and Heng, Jeremy and Doucet, Arnaud},
  journal={Advances in Neural Information Processing Systems},
  volume={34},
  pages={17695--17709},
  year={2021}
}

@article{saharia2022image,
  title={Image super-resolution via iterative refinement},
  author={Saharia, Chitwan and Ho, Jonathan and Chan, William and Salimans, Tim and Fleet, David J and Norouzi, Mohammad},
  journal={IEEE transactions on pattern analysis and machine intelligence},
  volume={45},
  number={4},
  pages={4713--4726},
  year={2022},
  publisher={IEEE}
}

@article{cao2024diffusion,
  title={Diffusion model with disentangled modulations for sharpening multispectral and hyperspectral images},
  author={Cao, Zihan and Cao, Shiqi and Deng, Liang-Jian and Wu, Xiao and Hou, Junming and Vivone, Gemine},
  journal={Information Fusion},
  volume={104},
  pages={102158},
  year={2024},
  publisher={Elsevier}
}

@article{meng2023pandiff,
  title={PanDiff: A novel pansharpening method based on denoising diffusion probabilistic model},
  author={Meng, Qingyan and Shi, Wenxu and Li, Sijia and Zhang, Linlin},
  journal={IEEE Transactions on Geoscience and Remote Sensing},
  volume={61},
  pages={1--17},
  year={2023},
  publisher={IEEE}
}

@article{rui2024unsupervised,
  title={Unsupervised hyperspectral pansharpening via low-rank diffusion model},
  author={Rui, Xiangyu and Cao, Xiangyong and Pang, Li and Zhu, Zeyu and Yue, Zongsheng and Meng, Deyu},
  journal={Information Fusion},
  volume={107},
  pages={102325},
  year={2024},
  publisher={Elsevier}
}

@article{vivone2019robust,
  title={Robust band-dependent spatial-detail approaches for panchromatic sharpening},
  author={Vivone, Gemine},
  journal={IEEE transactions on Geoscience and Remote Sensing},
  volume={57},
  number={9},
  pages={6421--6433},
  year={2019},
  publisher={IEEE}
}

@article{aiazzi2006mtf,
  title={MTF-tailored multiscale fusion of high-resolution MS and Pan imagery},
  author={Aiazzi, Bruno and Alparone, Luciano and Baronti, Stefano and Garzelli, Andrea and Selva, Massimo},
  journal={Photogrammetric Engineering \& Remote Sensing},
  volume={72},
  number={5},
  pages={591--596},
  year={2006},
  publisher={American Society for Photogrammetry and Remote Sensing}
}

@inproceedings{wei2017multi,
  title={Multi-scale-and-depth convolutional neural network for remote sensed imagery pan-sharpening},
  author={Wei, Yancong and Yuan, Qiangqiang and Meng, Xiangchao and Shen, Huanfeng and Zhang, Liangpei and Ng, Michael},
  booktitle={2017 IEEE International Geoscience and Remote Sensing Symposium (IGARSS)},
  pages={3413--3416},
  year={2017},
  organization={IEEE}
}

@inproceedings{zhou2022pan,
  title={Pan-sharpening with customized transformer and invertible neural network},
  author={Zhou, Man and Huang, Jie and Fang, Yanchi and Fu, Xueyang and Liu, Aiping},
  booktitle={Proceedings of the AAAI conference on artificial intelligence},
  volume={36},
  number={3},
  pages={3553--3561},
  year={2022}
}

@article{zhou2023memory,
  title={Memory-augmented deep unfolding network for guided image super-resolution},
  author={Zhou, Man and Yan, Keyu and Pan, Jinshan and Ren, Wenqi and Xie, Qi and Cao, Xiangyong},
  journal={International Journal of Computer Vision},
  volume={131},
  number={1},
  pages={215--242},
  year={2023},
  publisher={Springer}
}

@inproceedings{peng2022source,
  title={Source-Adaptive Discriminative Kernels based Network for Remote Sensing Pansharpening.},
  author={Peng, Siran and Deng, Liang-Jian and Hu, Jin-Fan and Zhuo, Yu-Wei},
  booktitle={IJCAI},
  pages={1283--1289},
  year={2022}
}

@inproceedings{shen2021efficient,
  title={Efficient attention: Attention with linear complexities},
  author={Shen, Zhuoran and Zhang, Mingyuan and Zhao, Haiyu and Yi, Shuai and Li, Hongsheng},
  booktitle={Proceedings of the IEEE/CVF winter conference on applications of computer vision},
  pages={3531--3539},
  year={2021}
}

@inproceedings{katharopoulos2020transformers,
  title={Transformers are rnns: Fast autoregressive transformers with linear attention},
  author={Katharopoulos, Angelos and Vyas, Apoorv and Pappas, Nikolaos and Fleuret, Fran{\c{c}}ois},
  booktitle={International conference on machine learning},
  pages={5156--5165},
  year={2020},
  organization={PMLR}
}

@article{vaswani2017attention,
  title={Attention is all you need},
  author={Vaswani, Ashish and Shazeer, Noam and Parmar, Niki and Uszkoreit, Jakob and Jones, Llion and Gomez, Aidan N and Kaiser, {\L}ukasz and Polosukhin, Illia},
  journal={Advances in neural information processing systems},
  volume={30},
  year={2017}
}

@article{tucker1966some,
  title={Some mathematical notes on three-mode factor analysis},
  author={Tucker, Ledyard R},
  journal={Psychometrika},
  volume={31},
  number={3},
  pages={279--311},
  year={1966},
  publisher={Springer}
}

@article{kolda2009tensor,
  title={Tensor decompositions and applications},
  author={Kolda, Tamara G and Bader, Brett W},
  journal={SIAM review},
  volume={51},
  number={3},
  pages={455--500},
  year={2009},
  publisher={SIAM}
}

@inproceedings{ronneberger2015u,
  title={U-net: Convolutional networks for biomedical image segmentation},
  author={Ronneberger, Olaf and Fischer, Philipp and Brox, Thomas},
  booktitle={Medical image computing and computer-assisted intervention--MICCAI 2015: 18th international conference, Munich, Germany, October 5-9, 2015, proceedings, part III 18},
  pages={234--241},
  year={2015},
  organization={Springer}
}

@inproceedings{liu2024residual,
  title={Residual denoising diffusion models},
  author={Liu, Jiawei and Wang, Qiang and Fan, Huijie and Wang, Yinong and Tang, Yandong and Qu, Liangqiong},
  booktitle={Proceedings of the IEEE/CVF Conference on Computer Vision and Pattern Recognition},
  pages={2773--2783},
  year={2024}
}

@article{he2023hqg,
  title={Hqg-net: Unpaired medical image enhancement with high-quality guidance},
  author={He, Chunming and Li, Kai and Xu, Guoxia and Yan, Jiangpeng and Tang, Longxiang and Zhang, Yulun and Wang, Yaowei and Li, Xiu},
  journal={IEEE Transactions on Neural Networks and Learning Systems},
  year={2023},
  publisher={IEEE}
}

@inproceedings{
hu2022lora,
title={Lo{RA}: Low-Rank Adaptation of Large Language Models},
author={Edward J Hu and Yelong Shen and Phillip Wallis and Zeyuan Allen-Zhu and Yuanzhi Li and Shean Wang and Lu Wang and Weizhu Chen},
booktitle={International Conference on Learning Representations},
year={2022},
url={https://openreview.net/forum?id=nZeVKeeFYf9}
}

@article{he2025pan,
  title={Pan-mamba: Effective pan-sharpening with state space model},
  author={He, Xuanhua and Cao, Ke and Zhang, Jie and Yan, Keyu and Wang, Yingying and Li, Rui and Xie, Chengjun and Hong, Danfeng and Zhou, Man},
  journal={Information Fusion},
  volume={115},
  pages={102779},
  year={2025},
  publisher={Elsevier}
}

@article{li2025diffusion,
  title={Diffusion models for image restoration and enhancement: a comprehensive survey},
  author={Li, Xin and Ren, Yulin and Jin, Xin and Lan, Cuiling and Wang, Xingrui and Zeng, Wenjun and Wang, Xinchao and Chen, Zhibo},
  journal={International Journal of Computer Vision},
  pages={1--31},
  year={2025},
  publisher={Springer}
}

@InProceedings{Xiao_2025_CVPR,
    author    = {Xiao, Jin-Liang and Huang, Ting-Zhu and Deng, Liang-Jian and Lin, Guang and Cao, Zihan and Li, Chao and Zhao, Qibin},
    title     = {Hyperspectral Pansharpening via Diffusion Models with Iteratively Zero-Shot Guidance},
    booktitle = {Proceedings of the IEEE/CVF Conference on Computer Vision and Pattern Recognition (CVPR)},
    month     = {June},
    year      = {2025},
    pages     = {12669-12678}
}

@article{yang2019deep,
  title={Deep learning for single image super-resolution: A brief review},
  author={Yang, Wenming and Zhang, Xuechen and Tian, Yapeng and Wang, Wei and Xue, Jing-Hao and Liao, Qingmin},
  journal={IEEE Transactions on Multimedia},
  volume={21},
  number={12},
  pages={3106--3121},
  year={2019},
  publisher={IEEE}
}

@inproceedings{he2016deep,
  title={Deep residual learning for image recognition},
  author={He, Kaiming and Zhang, Xiangyu and Ren, Shaoqing and Sun, Jian},
  booktitle={Proceedings of the IEEE conference on computer vision and pattern recognition},
  pages={770--778},
  year={2016}
}

@inproceedings{nichol2021improved,
  title={Improved denoising diffusion probabilistic models},
  author={Nichol, Alexander Quinn and Dhariwal, Prafulla},
  booktitle={International conference on machine learning},
  pages={8162--8171},
  year={2021},
  organization={PMLR}
}

@article{hendrycks2016gaussian,
  title={Gaussian error linear units (gelus)},
  author={Hendrycks, Dan and Gimpel, Kevin},
  journal={arXiv preprint arXiv:1606.08415},
  year={2016}
}
